%% file: main.tex
\documentclass[10pt,twocolumn,letterpaper]{article}
\usepackage{marvosym}
\usepackage{cvpr}      
\usepackage{multirow}
\input{preamble}

\definecolor{cvprblue}{rgb}{0.21,0.49,0.74}
\usepackage[pagebackref,breaklinks,colorlinks,citecolor=cvprblue]{hyperref}
\usepackage{colortbl}
\usepackage[dvipsnames]{xcolor}
\usepackage{makecell}
\usepackage{rotating}
\def\paperID{5000} 
\def\confName{CVPR}
\def\confYear{2024}

\title{Boosting Continual Learning of Vision-Language Models via Mixture-of-Experts Adapters}

\author{
    {Jiazuo Yu$^{1}$, Yunzhi Zhuge$^{1}$, Lu Zhang$^{1,}$\thanks{Corresponding author} , Ping Hu$^{2}$, Dong Wang$^{1}$, Huchuan Lu$^{1}$ and You He$^{3}$}\\
    {$^1$ Dalian University of Technology, China}\\
    {$^2$ University of Electronic Science and Technology of China}\\
    {$^3$ Tsinghua University, China}\\
    {\tt\small yujiazuo@mail.dlut.edu.cn, zhangluu@dlut.edu.cn} \\
    }

\begin{document}
\maketitle
\input{sec/0_abstract}    
\input{sec/1_intro}

\input{sec/2_related}

\input{sec/3_method}

\input{sec/4_experiments}
\input{sec/5_conclusion}

{
    \small
    \bibliographystyle{ieeenat_fullname}
    \bibliography{main}
}
\input{X_suppl}

\end{document}

%% file: preamble.tex
\usepackage[dvipsnames]{xcolor}


%% file: sec/0_abstract.tex
\begin{abstract}
Continual learning can empower vision-language models to continuously acquire new knowledge, without the need for access to the entire historical dataset. However, mitigating the performance degradation in large-scale models is non-trivial due to (i) parameter shifts throughout lifelong learning and (ii) significant computational burdens associated with full-model tuning. In this work, we present a parameter-efficient continual learning framework to alleviate long-term forgetting in incremental learning with vision-language models. Our approach involves the dynamic expansion of a pre-trained CLIP model, through the integration of Mixture-of-Experts (MoE) adapters in response to new tasks. To preserve the zero-shot recognition capability of vision-language models, we further introduce a Distribution Discriminative Auto-Selector (DDAS) that automatically routes in-distribution and out-of-distribution inputs to the MoE Adapter and the original CLIP, respectively. Through extensive experiments across various settings, our proposed method consistently outperforms previous state-of-the-art approaches while concurrently reducing parameter training burdens by 60\%. Our code locates at \url{https://github.com/JiazuoYu/MoE-Adapters4CL}
\end{abstract}

%% file: sec/1_intro.tex
\begin{figure*}[t]
	\centering
	\includegraphics[width=0.9\linewidth]{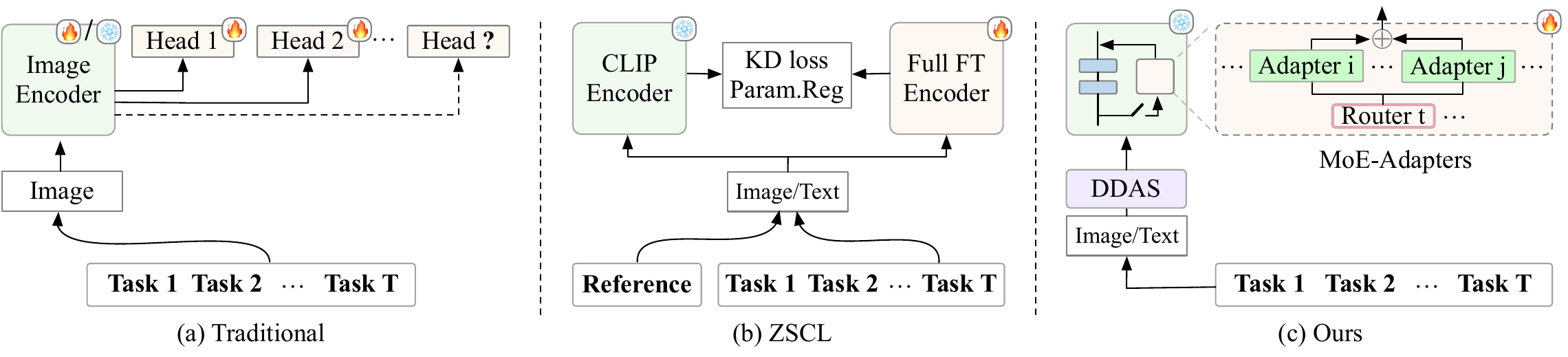}
	\vspace{-4pt}
	\caption{Comparison of various popular architectures to address CL. (a) Traditional dynamic expansion-based CL cannot distinguish unseen data. (b) Zero-shot CL~\cite{zheng2023preventing} suffers from significant computational burdens. (c) The proposed MoE-Adapters and DDAS collaborate to form a parameter-efficient, zero-shot CL.}
	\label{fig:teaser}
 \vspace{-10pt}
\end{figure*}

\section{Introduction}
\label{sec:intro}

Artificial Intelligence (AI), particularly in the realm of large-scale foundation models, has made significant strides in understanding the open world, as evidenced by recent advancements~\cite{OpenAI_2023, touvron2023llama, radford2021learning, liu2023visual, liu2023grounding}. An ideal AI, akin to human cognition, should be able to continuously assimilate new knowledge from the dynamic environment. Traditional fully-supervised training paradigms can't adapt to this scenario due to the high computational costs of integrating new data with historical datasets. In contrast, Continual Learning (CL), offering an efficient incremental training strategy, emerges as a solution by focusing on new data at each training stage. However, CL faces the significant hurdle of ``catastrophic forgetting'' where a model loses previously acquired knowledge upon learning new tasks~\cite{mccloskey1989catastrophic, goodfellow2013empirical}.

To remedy this issue, one of the popular solutions in current CL methods~\cite{aljundi2017expert, donahue2014decaf, girshick2014rich, li2017learning} is to develop dynamic expansion frameworks by incrementally adding task-specific components to a shared base model (see Figure~\ref{fig:teaser} (a)). 
Although these methods show promise in memorization and scalability, they cannot distinguish unseen data and thus overlook zero-shot transfer capability.
Recent advancements like ZSCL~\cite{zheng2023preventing} have brought the zero-shot transfer ability into continual learning by leveraging a pretrained Vision Language Model (VLM). As illustrated in Figure~\ref{fig:teaser} (b), this method relies on knowledge distillation to integrate zero-shot generalization ability from the frozen CLIP and uses parameter regularization to prevent knowledge degradation in continual learning. 
However, these designs often entail large computational burdens and exhibit limitations in long-term memorization. It's then natural to ask whether we can combine the merits of the pretrained foundation model and dynamic expansion strategy to form an effective system with robust memorization and zero-shot transfer abilities.

Recently, Parameter-Efficient Fine-Tune (PEFT) methods \cite{gao2023clip, sung2022vl, zhang2021tip, zaken2021bitfit, hu2021lora} have demonstrated that large-scale models can quickly adapt to downstream tasks via only fine-tuning less-parameterized adapters. This inspires us to build a dynamic expansion framework on VLM with task-specific adapters to relieve the parameter burdens in long-term CL.  Nevertheless, the intuitive approach of stacking adapters during incremental learning introduces a dependency on task identity. This poses challenges in practical scenarios such as class incremental learning where task identity may be unavailable. Furthermore, the use of independent adapters neglects the potential for inter-task knowledge sharing and cooperation, resulting in a limited representation capability and efficacy.

To overcome the outlined challenges, we propose a parameter-efficient continual learning framework by leveraging the recent advance in the field of multi-task learning, Mixture-of-Experts (MoE)~\cite{jacobs1991adaptive, shazeer2017outrageously}. We build a dynamic expansion architecture on a frozen CLIP model~\cite{radford2021learning}, dubbed as incremental MoE-Adapters, in which we take adapters as sparse experts and utilize incrementally incorporated task-specific routers to select the corresponding experts. In the continual learning process, we further apply a novel activate-freeze strategy to help the experts learn intra-task knowledge and encourage inter-task collaboration. Additionally, a Distribution Discriminative Auto-Selector (DDAS) is proposed to automatically allocate the testing data to MoE-Adapters or the pretrained CLIP, enabling effective predictions for seen data and zero-shot transfer for unseen data within a unified framework.

Our extensive experiments across various settings demonstrate the proposed method's effectiveness in addressing the catastrophic forgetting issue, significantly reducing the $60\%$ parameter burdens and memory requirements during training. Furthermore, when applied to few-shot continual learning, the proposed model shows exceptional resistance to forgetting and outperforms the previous arts by $3.6\%$, $7.0\%$ and 4.2\% in a 5-shot setting. 
Our contributions can be summarized as follows:
\begin{itemize}
\item We introduce a parameter-efficient training framework for vision-language models in continual learning, employing a MoE-Adapters based dynamic expansion architecture for enhanced adaptability and efficiency.
\item We develop an incremental activate-freeze strategy in the MoE framework, enabling experts to simultaneously acquire intra-task knowledge and engage in inter-task collaboration.
\item We design a Distribution Discriminative Auto-Selector (DDAS) for automated substream assignment, effectively merging anti-forgetting and zero-shot transfer capabilities within a unified model.
\end{itemize}

%% file: sec/2_related.tex
\section{Related Works}
\label{sec:Related Works}
\textbf{Continual Learning.}
Depending on the domain variations of incremental data, existing continual learning methods mainly focus on addressing \textit{i.e.,} Class Incremental Learning (CIL)~\cite{de2023effectiveness, jodelet2023class, belouadah2019il2m, yan2021dynamically, liu2023class} and Task Incremental Learning (TIL)~\cite{oren2021defense, mallya2018packnet, zheng2023preventing}. 
Existing efforts in this area have been made by developing various architectures~\cite{de2021continual}, including memory-based, regularization-based and dynamic-based models.  
Memory-based methods~\cite{shin2017continual, lopez2017gradient, prabhu2020gdumb, lavda2018continual, ostapenko2022continual, rebuffi2017icarl, isele2018selective} retain the historical knowledge by storing them in a memory bank, which will be accessed and updated in incremental learning. However, the continuously increasing learned data usually poses a burden on the memory bank, resulting in limited lifelong learning ability. 
Regularization-based methods add explicit regularization terms on weights~\cite{aljundi2018memory, kirkpatrick2017overcoming, lee2017overcoming, zenke2017continual} or data~\cite{dhar2019learning, douillard2020podnet, hou2019learning, li2017learning} to balance between the older and new tasks. They are usually used as an auxiliary trick in memory-based or dynamic models to alleviate the forgetting issue.
Dynamic methods~\cite{douillard2022dytox, hu2023dense, ye2023self, yan2021dynamically, yoon2017lifelong, aljundi2017expert} address continual learning by incrementally adding new parameters on the baseline, such as neurons, branches or prediction heads. Dynamic methods usually perform favorably against the other two pipelines. However, like memory-based methods, the dynamic architecture often incurs large-scale model sizes, limiting the models' efficiency.
Despite the promising performance of the approaches aforementioned, addressing the crucial capability of AI agents, namely zero-shot transfer to unseen knowledge, remains challenging and complex to integrate into existing popular pipelines. In this paper, we propose incorporating the dynamic architecture on vision-language models to boost their memorization of historical knowledge and alleviate the degradation of zero-shot transfer ability.        
The highly related work is ZSCL~\cite{zheng2023preventing}, which uses parameter regularization in the continual learning of large-scale models. In contrast to the fully finetuning strategy in ZSCL, our method proposed an incremental MoE adapter to decrease the tuned parameters and enhance the collaboration of historically learned adapters and ongoing ones.


\begin{figure*}[t]
	\centering
	\includegraphics[width=0.98\linewidth]{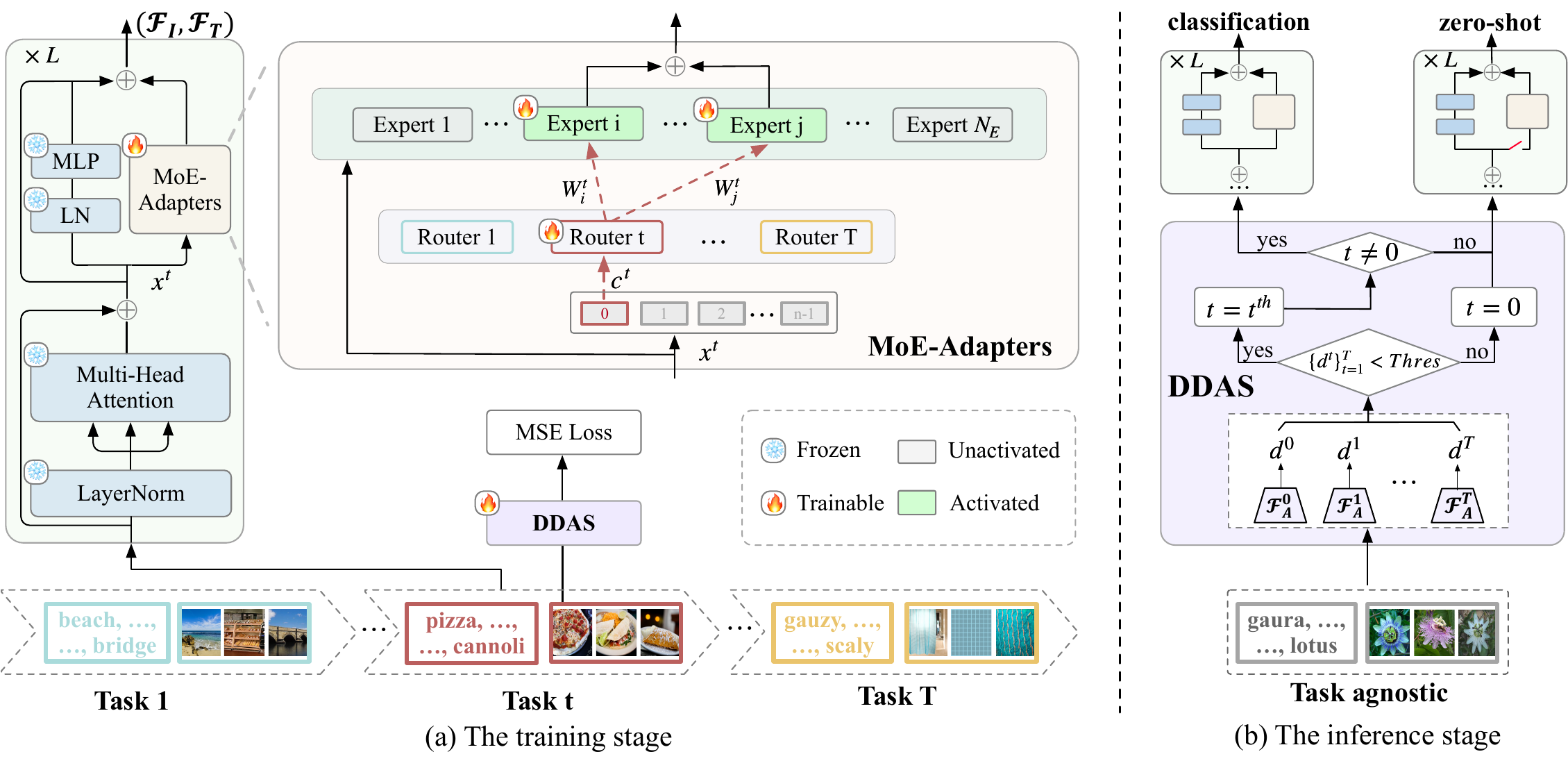}
 \vspace{-4pt}
        \caption{Overall framework of the proposed method. (a) At the training stage, CLIP's image and text encoders $(\mathcal{F}_I,\mathcal{F}_T)$ take input samples from \textbf{Task \textit{t}}. 
        In each of transformer blocks, there is a MoE-Adapters, whose input is the tokens $ \textbf{\textit{x}}^t $ from MHSA. The router takes the task-specific \texttt{[CLS]} token $ \textbf{\textit{c}}^t $ as input and produces experts' weights $W_i^t$ and $W_j^t$ to combine the expert's output. DDAS is trained using only images via the MSE loss defined by Eq.~\ref{eq:chooseid1}. (b) At the inference stage, the proposed DDAS determines the data distribution by comparing the distribution $\{d^t\}_{t=1}^T$ in each autoencoder of the \textbf{task-agnostic} images. It can automatically assign the testing data into MoE-Adapters or original CLIP to predict with either seen or unseen data. 
        }
	\label{fig:overview}
 \vspace{-9pt}
\end{figure*}
\noindent\textbf{Parameter Efficient Fine-Tuning.}
In the realm of Natural Language Processing (NLP), fine-tuning large-scale models (\emph{e.g.,} 175B GPT-3~\cite{brown2020language}) imposes significant burdens in both parameter complexity and time consumption. Thus, several parameter-efficient fine-tuning methods~\cite{zhang2020side, hu2021lora, jia2022visual, houlsby2019parameter, karimi2021compacter} have been explored, which only set a few trainable parameters and fine-tune them for efficiency. Among these methods, LoRA \cite{hu2021lora} and Compacter~\cite{karimi2021compacter} reduce the number of trainable parameters by attaching low-rank hyper-complex adapter layers or sharing adapter parameters across layers, respectively.
The success of efficient tuning strategies in NLP promotes their applications on vision-language models~\cite{ju2022prompting, zhou2022conditional, gao2023clip, sung2022vl, zhang2021tip} like CLIP~\cite{gao2023clip}.
Recently, Liu \textit{et al.}~\cite{liu2023class} introduce efficient tuning strategies in the continual learning of CLIP, which uses trainable adapters and a parameter retention strategy for downstream task adaptation and historical knowledge memorization, respectively.
However, this method is only applied to CIL and ignores the zero-shot ability of the original CLIP. 
In this paper, we propose a novel parameter-efficient tuning approach on CLIP to boost both the anti-forgetting and zero-shot abilities in continual learning. Our model can flexibly adapt to CIL and TIL and achieve promising performance even trained with few data (namely few-shot continual learning).

\noindent\textbf{Mixture of Experts.}
The MoE~\cite{jacobs1991adaptive} contains multiple experts and a routing network. It aggregates the expert outputs via a weighted strategy by the routing network. Based on the sparse architecture of MoE~\cite{shazeer2017outrageously}, some methods~\cite{riquelme2021scaling, mustafa2022multimodal,fedus2022switch, daxberger2023mobile} are proposed to decrease computational costs and improve model capacity. This technique is also introduced to continual learning to mitigate the forgetting issue. For example, Aljundi \textit{et al.}~\cite{aljundi2017expert} propose to train multiple backbones as experts and automatically feed the test samples to a relevant expert. Chen \textit{et al.}~\cite{chen2023lifelong} utilize the pre-trained experts and gates to store previous knowledge. These methods have demonstrated MoE's promising performance in continual learning.
We propose an incremental MoE-Adapters for continual learning with CLIP. We use adapters as experts to increase the adaption speed and introduce an incremental expert interaction strategy to facilitate the collaboration of experts during continual learning.

%% file: sec/3_method.tex
\section{Methodology}

\subsection{Continual Learning}
Given a set of $T$ tasks $\{{\mathcal{T}^t}\}_{t=1}^{T}$, continual learning works by sequentially accessing and learning on each task $\mathcal{T}^t=\{\mathcal{D}^t,\mathcal{C}^t\}$. Here, $\mathcal{D}^t=\{{I}^{t}_i,{y}^{t}_i\}_{i=1}^{N^t}$ represents the data of $t^{th}$ task $\mathcal{T}^t$, where $I^{t}_{i}$ is the input image, $y^{t}_{i} \in \mathcal{C}^t$ is the corresponding class label, and $N^t$ is the size of data. The category set $\mathcal{C}^t=\{c^{t}_j\}_{j=1}^{M^t}$ encompasses the class names within $\mathcal{T}^t$, with a total of $M^t$ classes.
Continual learning aims to achieve good performance across all tasks and can be broadly categorized into Task Incremental Learning (TIL) and Class Incremental Learning (CIL).  In TIL, the model generates predictions within a task-specific set $\mathcal{C}^t$, which is determined by the current task identity $t$. Meanwhile, in CIL, the challenge involves distinguishing between all the previously encountered classes $\cup_{i=1}^t \mathcal{C}^i$.


\subsection{Framework Overview}

In this paper, we present a parameter-efficient framework designed to empower the continual learning capabilities of vision-language models, achieving robust historical knowledge memorization without sacrificing the  zero-shot generalization abilities. Our method is built upon the CLIP~\cite{gao2023clip} model, which contains parallel encoders $(\mathcal{F}_I,\mathcal{F}_T)$ to extract features of input images and texts, respectively. By following  CLIP~\cite{gao2023clip}, we make predictions based on the cosine similarity between the final image embedding $\mathcal{F}_I(I_i^t)$ and text embedding $\mathcal{F}_T(c^{t}_{j})$. The input image is then assigned to the class with the highest similarity.

The overall framework of our method is shown in Figure~\ref{fig:overview}. We introduce MoE structure onto a frozen CLIP to consolidate all the downstream tasks within a unified model, in which the task-dependent routers are sequentially added to modulate the experts for each task. Adapter modules, such as LoRA~\cite{hu2021lora}, function as the experts in the MoE setup, enhancing adaptation speed during training. To relieve MoE's reliance  on task identities, we further propose a Distribution Discriminative Auto-Selector (DDAS). The DDAS automatically infers the task context by analyzing the variations of the target image distributions.  As a result, in-distribution data will be allocated to the corresponding routers within MoE, while the out-of-distribution inputs will be identified and directed  to the original CLIP to perform zero-shot recognition.    


\subsection{Incremental Mixture-of-Experts Adapters}
\label{MoEs}

We leverage MoE~\cite{shazeer2017outrageously} to build an expansible architecture to alleviate the ``catastrophic forgetting'' issue in the continual learning of CLIP. The MoE is composed of several experts $\{\mathcal{E}_i\}_{i=1}^{N_E}$ and routers, where $N_E$ is the number of predefined experts. For current task $\mathcal{T}^t$, only a task-dependent router $\mathcal{R}^t, t\in[1,T]$ is added to the system, which integrates the experts' outputs via gated average.

\vspace{1ex}
\noindent\textbf{Adapters as Experts.} 
MoE in vision-language models usually incorporate the experts inside networks, which can be MLPs or attention heads~\cite{shazeer2017outrageously, zhang2022mixture, chen2023mod}. However, inserting the MoE inside VLM might bring in significant computational burdens due to the full-parameter tuning. Some methods~\cite{gao2023clip, liu2023class} have demonstrated that adapters with few parameters can increase the adaption speed of VLM on downstream tasks and enable their applications in continual learning. 
Inspired by this, we use the effective adapter LoRA~\cite{hu2021lora}, which works by decoupling the original heavy and frozen parameters into low-rank trainable space, as the experts in MoE to speed up continual learning with CLIP. 
%

Our MoE-Adapters are implemented in all the Transformer blocks of the parallel encoders $(\mathcal{F}_I,\mathcal{F}_T)$, as shown in Figure~\ref{fig:overview}. 
To be more specific, in each Transformer block, the feature tokens \textbf{\textit{$x^t$}} $\in \mathbb{R}^{n\times d}$ after the multi-head attention output are passed to all the experts in the MoE. 
Then, the task-specific router $\mathcal{R}^t$ is applied to fuse the experts' outputs via gated summation. Note that we implement the same MoE adapters in both the image  encoder and the textual encoder, without sharing the parameters. 
 
%

\vspace{1ex}
\noindent\textbf{Incremental Mixture of Experts.} 
In our MoE framework, task-specific routers $\mathcal{R}^t$ determine the activation of experts $\mathcal{E}_i$ to produce outcomes tailored to each task $t$. The combined output for a task, $\textbf{\textit{y}}^t$, is computed as:
\begin{equation}
    \textbf{\textit{y}}^t = \sum_{i=1}^{N_E}{W_i^t}\mathcal{E}_i(\textbf{\textit{x}}^t),
\end{equation}
where $W^t=\{W_i^t\}_{i=1}^{N_E}$ represents the gating weights assigned by $\mathcal{R}^t$, dictating each expert's contribution. $\textbf{\textit{x}}^t$ denotes the tokens processed for task $t$, and $\textbf{\textit{y}}^t$ is the corresponding output from the MoE-Adapters, matching the shape of $\textbf{\textit{x}}^t$.
We refine the MoE-Adapters for continual learning with two key modifications. Unlike previous methods~\cite{riquelme2021scaling, daxberger2023mobile} that input patch or image tokens into the router, we utilize the initial token, known as the \texttt{[CLS]} token ($\textbf{\textit{c}}^t \in \mathbb{R}^{1\times d}$), to enhance processing efficiency. The gating weights are then computed as follows:
\begin{equation}
\label{eq:gate1}
W^t = Softmax(Topk(\mathcal{R}^t(\textbf{\textit{c}}^t))),
\end{equation}
where $\mathcal{R}^t$ projects $\textbf{\textit{c}}^t$ to a 1-D vector indicating each expert's likelihood of activation. The $Topk(\cdot)$ function selects the $k$ most relevant experts, while setting the rest to be $-\infty$. The $Softmax(\cdot)$ function normalizes these weights to emphasize the selected experts' contribution.

\begin{figure}[t]
	\centering
	\includegraphics[width=1\linewidth]{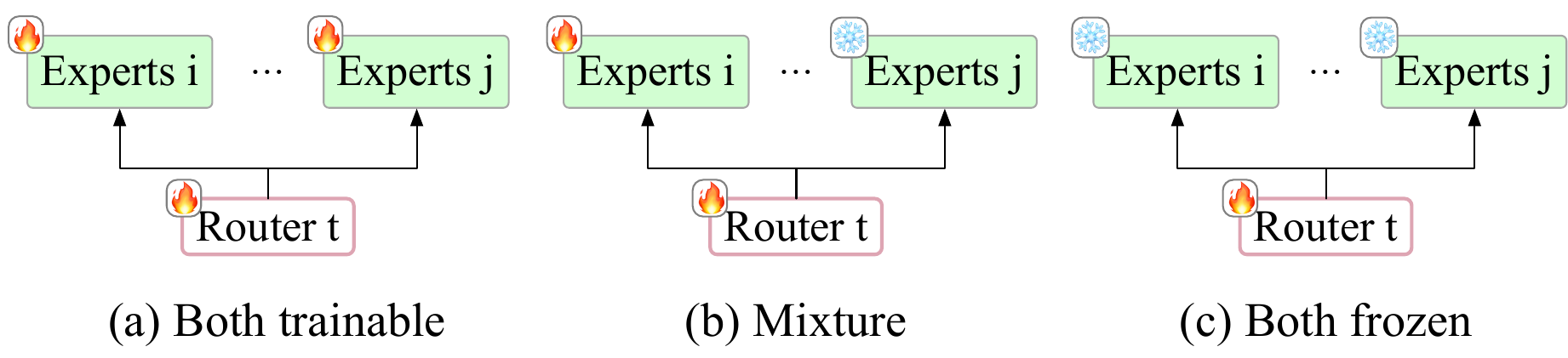}
	\vspace{-14pt}
	\caption{The three distinct combinations among activated experts (a) both trained, (b) trainable and frozen, (c) both experts are frozen, and only the router is trainable.}
	\label{fig:different frozen}
  \vspace{-8pt}
 \end{figure}

\begin{table*}
    \centering
    \resizebox{0.95\linewidth}{!}{
    \begin{tabular}{c>{\raggedright\arraybackslash}p{3cm} >{\centering\arraybackslash}p{1cm} >{\centering\arraybackslash}p{1cm}>{\centering\arraybackslash}p{1cm} >{\centering\arraybackslash}p{1cm} >{\centering\arraybackslash}p{1cm}>{\centering\arraybackslash}p{1cm} >{\centering\arraybackslash}p{1cm} >{\centering\arraybackslash}p{1cm}>{\centering\arraybackslash}p{1cm} >{\centering\arraybackslash}p{1cm} >{\centering\arraybackslash}p{1cm} >{\centering\arraybackslash}p{1.5cm}}
        \toprule
           & {\hspace{1em}} Method & \makecell[c]{\rotatebox{90}{Aircraft~\cite{maji2013fine}}} & \makecell[c]{\rotatebox{90}{Caltech101~\cite{fei2004learning}}} & \makecell[c]{\rotatebox{90}{CIFAR100~\cite{krizhevsky2009learning}}} & \makecell[c]{\rotatebox{90}{DTD~\cite{cimpoi2014describing}}} & \makecell[c]{\rotatebox{90}{EuroSAT~\cite{helber2019eurosat}}} & \makecell[c]{\rotatebox{90}{Flowers~\cite{nilsback2008automated}}} & \makecell[c]{\rotatebox{90}{Food~\cite{bossard2014food}}} & \makecell[c]{\rotatebox{90}{MNIST~\cite{deng2012mnist}}} & \makecell[c]{\rotatebox{90}{OxfordPet~\cite{parkhi2012cats}}} & \makecell[c]{\rotatebox{90}{Cars~\cite{krause20133d}}} & \makecell[c]{\rotatebox{90}{SUN397~\cite{xiao2010sun}}} & \makecell[c]{{\textit{Average}}} \\
  
        \midrule
        
            \multirow{3}{*}{\rotatebox{90}{CLIP}}& {\hspace{1em}}Zero-shot & 24.3 & 88.4 & 68.2 & 44.6 & 54.9 & 71.0 & 88.5 & 59.4 & 89.0 & 64.7 & 65.2 & 65.3  \\
            & {\hspace{1em}}Full Fine-tune & 62.0 & 95.1 & 89.6 & 79.5 & 98.9 & 97.5 & 92.7 & 99.6 & 94.7 & 89.6 & 81.8 & 89.2  \\
            & {\hspace{1em}}Fine-tune Adapter & 56.8 & 92.6 & 89.4 & 79.0 & 98.4 & 97.0 & 92.9 & 99.2 & 94.1 & 89.1 & 82.7 & 88.3 \\
            
            \midrule\midrule
            \multirow{8}{*}{\rotatebox{90}{\textbf{Transfer}}} &{\hspace{1em}}Continual-FT & & 67.1 & 46.0 & 32.1 & 35.6 & 35.0 & 57.7 & 44.1 & 60.8 & 20.5 & 46.6 & 44.6 \\
            & {\hspace{1em}}LwF \cite{li2017learning} &  & 74.5 & 56.9 & 39.1 & \textbf{51.1} & 52.6 & 72.8 & \underline{60.6} & 75.1 & 30.3 & 55.9 & 58.9 \\
            & {\hspace{1em}}iCaRL \cite{rebuffi2017icarl} &  & 56.6 & 44.6 & 32.7 & 39.3 & 46.6 & 68.0 & 46.0 & 77.4 & 31.9 & 60.5 & 50.4 \\
            & {\hspace{1em}}LwF-VR \cite{ding2022don} & & 77.1 & 61.0 & 40.5 & 45.3 & 54.4 & 74.6 & 47.9 & 76.7 & 36.3 & 58.6 & 57.2  \\
            & {\hspace{1em}}WiSE-FT \cite{wortsman2022robust} & & 73.5 & 55.6 & 35.6 & 41.5 & 47.0 & 68.3 & 53.9 & 69.3 & 26.8 & 51.9 & 52.3  \\
            & {\hspace{1em}}ZSCL \cite{zheng2023preventing} & & 86.0 & 67.4 & \textbf{45.4} & \underline{50.4} & \underline{69.1} & 87.6 & \textbf{61.8} & 86.8 & 60.1 & \textbf{66.8} & \underline{68.1} \\
            \rowcolor{Thistle!20}
            & {\hspace{1em}}Ours$\dagger$ & & \textbf{87.9} & \textbf{68.2} & 42.2 & 41.4 & 68.7 & \textbf{88.7} & 59.4 & \textbf{89.1} & \textbf{64.5} & 64.0 & 67.4\textbf{\textcolor{NavyBlue}{ (-0.7)}}  \\
            \rowcolor{Thistle!20}
            & {\hspace{1em}}Ours &&\textbf{87.9} & \textbf{68.2} & \underline{44.4} &49.9 &\textbf{70.7} & \textbf{88.7} &59.7 &\textbf{89.1} &\textbf{64.5} &\underline{65.5} &\textbf{68.9\textcolor{Maroon}{ (+0.8)}}\\
            \midrule 
            \multirow{8}{*}{\rotatebox{90}{\textbf{Average}}} &{\hspace{1em}}Continual-FT    & 25.5 & 81.5 & 59.1 & 53.2 & 64.7 & 51.8 & 63.2 & 64.3 & 69.7 & 31.8 & 49.7 & 55.9 \\
            & {\hspace{1em}}LwF \cite{li2017learning} & 36.3 & 86.9 & 72.0 & 59.0 & 73.7 & 60.0 & 73.6 & \underline{74.8} & 80.0 & 37.3 & 58.1 & 64.7 \\
            & {\hspace{1em}}iCaRL \cite{rebuffi2017icarl} & 35.5 & 89.2 & 72.2 & 60.6 & 68.8 & 70.0 & 78.2 & 62.3 & 81.8 & 41.2 & 62.5 & 65.7 \\
            & {\hspace{1em}}LwF-VR \cite{ding2022don} & 29.6 & 87.7 & 74.4 & 59.5 & 72.4 & 63.6 & 77.0 & 66.7 & 81.2 & 43.7 & 60.7 & 65.1  \\
            & {\hspace{1em}}WiSE-FT \cite{wortsman2022robust} & 26.7 & 86.5 & 64.3 & 57.1 & 65.7 & 58.7 & 71.1 & 70.5 & 75.8 & 36.9 & 54.6 & 60.7  \\
            & {\hspace{1em}}ZSCL \cite{zheng2023preventing} & 45.1 & \textbf{92.0} & 80.1 & 64.3 & \textbf{79.5} & 81.6 & \textbf{89.6} & \textbf{75.2} & 88.9 & 64.7 & \textbf{68.0} &75.4 \\
            \rowcolor{Thistle!20}
            & {\hspace{1em}}Ours$\dagger$& \textbf{54.3} & 91.1 & \textbf{85.1} & \textbf{69.7} & 77.5 & \textbf{84.5} & \underline{89.1} &  73.8 & \underline{89.2} & \textbf{69.0} & 65.8 & \textbf{77.2\textcolor{Maroon}{\textbf{(+1.8)}}} \\
            \rowcolor{Thistle!20}
            & {\hspace{1em}}Ours &\underline{50.2}&\underline{91.9}&\underline{83.1}&\underline{69.4}&\underline{78.9}&\underline{84.0}&\underline{89.1}&73.7&\textbf{89.3}&\underline{67.7}&\underline{66.9}&\underline{76.7}\textcolor{Maroon}{\textbf{(+1.3)}}\\
            \midrule
            \multirow{8}{*}{\rotatebox{90}{\textbf{Last}}} &{\hspace{1em}}Continual-FT & 31.0 & 89.3 & 65.8 & 67.3 & 88.9 & 71.1 & 85.6 & \textbf{99.6} & 92.9 & 77.3 & 81.1 & 77.3 \\
            & {\hspace{1em}}LwF \cite{li2017learning} & 26.3 & 87.5 & 71.9 & 66.6 & 79.9 & 66.9 & 83.8 & \textbf{99.6} & 92.1 & 66.1 & 80.4 & 74.6 \\
            & {\hspace{1em}}iCaRL \cite{rebuffi2017icarl} & 35.8 & \textbf{93.0} & 77.0 & 70.2 & 83.3 & 88.5 & \underline{90.4} & 86.7 & \underline{93.2} & 81.2 & \underline{81.9} & 80.1 \\
            & {\hspace{1em}}LwF-VR \cite{ding2022don} & 20.5 & 89.8 & 72.3 & 67.6 & 85.5 & 73.8 & 85.7 & \textbf{99.6} & 93.1 & 73.3 & 80.9 & 76.6  \\
            & {\hspace{1em}}WiSE-FT \cite{wortsman2022robust} & 27.2 & 90.8 & 68.0 & 68.9 & 86.9 & 74.0 & 87.6 & \textbf{99.6} & 92.6 & 77.8 & 81.3 & 77.7  \\
            & {\hspace{1em}}ZSCL \cite{zheng2023preventing} & 40.6 & \underline{92.2} & 81.3 & 70.5 & 94.8 & 90.5 & \textbf{91.9} & 98.7 & \textbf{93.9} & \underline{85.3} & 80.2 & 83.6 \\
            \rowcolor{Thistle!20}
            & {\hspace{1em}}Ours$\dagger$ & \textbf{54.3} & 90.8 & \textbf{88.8} & \textbf{80.3} & \textbf{98.1} & \textbf{97.5} & 89.6 & 99.1 & 89.5 & \textbf{89.2} & \textbf{83.8} & \textbf{87.4\textcolor{Maroon}{ (+3.8)}} \\ 
            \rowcolor{Thistle!20}
            & {\hspace{1em}}Ours& \underline{49.8}&\underline{92.2}&\underline{86.1}&\underline{78.1}&\underline{95.7}&\underline{94.3}&89.5&98.1&89.9&81.6&80.0&\underline{85.0}\textcolor{Maroon}{\textbf{(+1.4)}}\\
        \bottomrule
    \end{tabular}}
    \vspace{-4pt}
    \caption{Comparison with state-of-the-art methods on MTIL benchmark in terms of ``Transfer'', ``Average'', and ``Last'' scores (\%). ``Ours$\dagger$'' and ``Ours'' indicate our method trained on 3k and 1k iterations, respectively. We label the best and second methods with \textbf{bold} and \underline{underline} styles. The top block indicates the upper-bound solutions to adapt the CLIP on each task.}
    \label{tab:compareAvg_Last_Transfer}
     \vspace{-11pt}
\end{table*}

\begin{table*}[t]
	\centering
        \resizebox{0.99\linewidth}{!}{
	\begin{tabular}{cl>{\centering\arraybackslash}p{1cm} >{\centering\arraybackslash}p{1cm}>{\centering\arraybackslash}p{1cm} >{\centering\arraybackslash}p{1cm} >{\centering\arraybackslash}p{1cm} >{\centering\arraybackslash}p{1cm} >{\centering\arraybackslash}p{1cm} >{\centering\arraybackslash}p{1cm} >{\centering\arraybackslash}p{1cm} >{\centering\arraybackslash}p{1cm} >{\centering\arraybackslash}p{1cm} >{\centering\arraybackslash}p{1.5cm}}
 
		\toprule
               &{\hspace{1em}} Method & \makecell[c]{\rotatebox{90}{Aircraft~\cite{maji2013fine}}} & \makecell[c]{\rotatebox{90}{Caltech101~\cite{fei2004learning}}} & \makecell[c]{\rotatebox{90}{CIFAR100~\cite{krizhevsky2009learning}}} & \makecell[c]{\rotatebox{90}{DTD~\cite{cimpoi2014describing}}} & \makecell[c]{\rotatebox{90}{EuroSAT~\cite{helber2019eurosat}}} & \makecell[c]{\rotatebox{90}{Flowers~\cite{nilsback2008automated}}} & \makecell[c]{\rotatebox{90}{Food~\cite{bossard2014food}}} & \makecell[c]{\rotatebox{90}{MNIST~\cite{deng2012mnist}}} & \makecell[c]{\rotatebox{90}{OxfordPet~\cite{parkhi2012cats}}} & \makecell[c]{\rotatebox{90}{Cars~\cite{krause20133d}}} & \makecell[c]{\rotatebox{90}{SUN397~\cite{xiao2010sun}}} & \makecell[c]{\textit{Average}} \\
  
		\midrule
		\multirow{3}{*}{\rotatebox{90}{CLIP}}&{\hspace{1em}}Zero-shot & 24.3 & 88.4 & 68.2 & 44.6 & 54.9 & 71.0 & 88.5 & 59.4 & 89.0 & 64.7 & 65.2 & 65.3  \\
            &{\hspace{1em}}5-shot Full Fine-tune & 30.6 & 93.5 & 76.8 & 65.1 & 91.7 & 92.9 & 83.3 & 96.6 & 84.9 & 65.4 & 71.3 & 77.5 \\
            &{\hspace{1em}}5-shot Fine-tune Adapter & 29.7 & 90.0 & 75.3 & 63.9 & 81.1 & 94.2 & 87.8 & 90.4 & 89.0 & 68.2 & 72.5 & 76.6 \\
            
            \midrule\midrule
            \multirow{7}{*}{\rotatebox{90}{\textbf{Transfer}}}&{\hspace{1em}}Continual-FT & & 72.8  & 53.0 &36.4  &35.4  & 43.3 & 68.4 & 47.4 & 72.6 & 30.0 &52.7  & 51.2   \\
            &{\hspace{1em}}LwF \cite{li2017learning} &  & 72.1 & 49.2 & 35.9 & 44.5 & 41.1 & 66.6 & 50.5 & 69.0 & 19.0 & 51.7 & 50.0   \\
            &{\hspace{1em}}LwF-VR \cite{ding2022don} &  & 82.2 & 62.5 & 40.1 & 40.1 & 56.3 & 80.0 & 60.9 & 77.6 & 40.5 & 60.8 & 60.1   \\
            &{\hspace{1em}}WiSE-FT \cite{wortsman2022robust} &  & 77.6 & 60.0 & 41.3 & 39.4 & 53.0 & 76.6 & 58.1 & 75.5 & 37.3 & 58.2 & 57.7   \\
            &{\hspace{1em}}ZSCL \cite{zheng2023preventing} &  & \underline{84.0} & \underline{68.1} & \textbf{44.8} & \underline{46.8} & \underline{63.6} & \underline{84.9} & \underline{61.4} & \underline{81.4} & \underline{55.5} & \underline{62.2} & \underline{65.3} \\
            \rowcolor{Thistle!20}
            &{\hspace{1em}}Ours &  & \textbf{87.9} & \textbf{68.2} & \underline{44.1} & \textbf{48.1} & \textbf{64.7} & \textbf{88.8} & \textbf{69.0} & \textbf{89.1} & \textbf{64.5} & \textbf{65.1} & \textbf{68.9\textcolor{Maroon}{(+3.6)}} \\
            \midrule
            \multirow{7}{*}{\rotatebox{90}{\textbf{Average}}}&{\hspace{1em}}Continual-FT\hphantom{Space} \hphantom{Space} & 28.1 & 86.4 &59.1 &52.8 & 55.8 &62.0 &70.2 & 64.7 & 75.5 & 35.0 &54.0 & 58.5 \\
            &{\hspace{1em}}LwF \cite{li2017learning} &23.5 &77.4  &43.5 &41.7 &43.5  &52.2 &54.6 & 63.4 & 68.0& 21.3& 52.6&49.2\\
            &{\hspace{1em}}LwF-VR \cite{ding2022don} &24.9 & \underline{89.1} &64.2 &53.4 &54.3  &70.8 &79.2 &66.5  &79.2 & 44.1& 61.6&62.5\\
            &{\hspace{1em}}WiSE-FT \cite{wortsman2022robust} & \textbf{32.0} & 87.7 & 61.0 &\underline{55.8} & \underline{68.1} & 69.3&76.8 &\underline{71.5}  &77.6 &42.0 &59.3&63.7  \\
            &{\hspace{1em}}ZSCL \cite{zheng2023preventing} & 28.2& 88.6 &\underline{66.5} & 53.5&56.3  &\underline{73.4} &\underline{83.1} & 56.4 & \underline{82.4} & \underline{57.5}&\underline{62.9}&\underline{64.4} \\
            \rowcolor{Thistle!20}
            &{\hspace{1em}}Ours & \underline{30.0} & \textbf{89.6} & \textbf{73.9}& \textbf{58.7}& \textbf{69.3} &\textbf{79.3} & \textbf{88.1}& \textbf{76.5} & \textbf{89.1}& \textbf{65.3}&\textbf{65.8} & \textbf{71.4\textcolor{Maroon}{(+7.0)}} \\
            
            \midrule
            \multirow{7}{*}{\rotatebox{90}{\textbf{Last}}}&{\hspace{1em}}Continual-FT & 27.8 & 86.9 & 60.1 & 58.4 & 56.6 & 75.7 & 73.8 & \underline{93.1} & 82.5 & 57.0 & 66.8 & 67.1 \\
            &{\hspace{1em}}LwF \cite{li2017learning} & 22.1 & 58.2 & 17.9 & 32.1 & 28.1 & 66.7 & 46.0 & 84.3 & 64.1 & 31.5 & 60.1 & 46.5 \\
            &{\hspace{1em}}LwF-VR \cite{ding2022don} & 22.9 & \textbf{89.8} & 59.3 & 57.1 & 57.6 & 79.2 & 78.3 & 77.7 & 83.6 & 60.1 & 69.8 & 66.9  \\
            &{\hspace{1em}}WiSE-FT \cite{wortsman2022robust} & \textbf{30.8} & 88.9 & 59.6 & \underline{60.3} & \underline{80.9} & 81.7 & 77.1 & \textbf{94.9} & 83.2 & 62.8 & 70.0 & \underline{71.9}  \\
            &{\hspace{1em}}ZSCL \cite{zheng2023preventing} &26.8 & 88.5 & \underline{63.7} & 55.7 & 60.2 & \underline{82.1} & \underline{82.6} & 58.6 & \underline{85.9} & \underline{66.7} & \underline{70.4} & 67.4 \\

            \rowcolor{Thistle!20}
            &{\hspace{1em}}Ours & \underline{30.1} & \underline{89.3} &\textbf{74.9}  & \textbf{64.0} & \textbf{82.3} &  \textbf{89.4}&\textbf{87.1} & 89.0 & \textbf{89.1} &\textbf{69.5} &\textbf{72.5}  & \textbf{76.1\textcolor{Maroon}{(+4.2)}} \\ 
            
		\bottomrule
	\end{tabular}}
 \vspace{-4pt}
	\caption{Comparison with state-of-the-art methods on few-shot MTIL benchmark in terms of ``Transfer”, ``Average”, and ``Last” scores (\%). Ours converges in 500 iterations on few-shot. We label the best and second methods with \textbf{bold} and \underline{underline} styles. The top block indicates the upper-bound solutions to adapt the CLIP on each task.}
	\label{tab:comparefs}
  \vspace{-11pt}
\end{table*}

\vspace{1ex}
\noindent\textbf{Training MoE-Adapters.}
We train the MoE-Adapters through  simple back-propagation, orchestrated by an incremental activate-freeze strategy. The objective is to augment experts with intra-task knowledge and inter-task collaboration. 
Specifically, after training on an older task, we count the distribution of its router's outputs. 
The $Top$-$k$ most activated experts are then kept frozen during subsequent task training to preserve task-specific knowledge. 
In this manner,  when faced with a new task, the respective router is able to access frozen experts for leveraging shareable knowledge from historical tasks, and optimize unfrozen experts to acquire specific information for the new task.
As illustrated in Figure~\ref{fig:different frozen}, during training the router can activate (a) only the untapped experts, (b) both untapped and previously learned experts, and (c) only the learned experts from previous tasks. As a result, this strategy allows experts to consolidate their knowledge collaboratively, resembling the human brain's mechanism of reinforcing and connecting new information with existing memories.

%


\subsection{Distribution Discriminative Auto-Selector}
\label{DDAS}
The task-specific nature of the routers in our MoE-Adapters necessitates manual task identity to activate the appropriate router. Such manual intervention is not aligned with the automated and practical nature of Task Incremental Learning (TIL) and Class Incremental Learning (CIL), and restricts the inherent zero-shot generalization capability of the CLIP model. To address this limitation, we develop the Distribution Discriminative Auto-Selector (DDAS), which automatically selects the proper router with the task context inferred by analyzing the variation in the distribution of input.

DDAS extends the incremental MoE framework by introducing a series of task-specific autoencoders, $\{\mathcal{F}_A^t\}_{t=1}^T$, which are trained to independently capture the distribution characteristics for the tasks $\{{\mathcal{T}^t}\}_{t=1}^{T}$. The loss function employed for training the autoencoders is the Mean Squared Error (MSE), defined as:
\begin{equation}
     d^t= ||\textit{\textbf{f}}^t_i - \textit{\textbf{f}}^t_o||^2,
  \label{eq:chooseid1}
\end{equation}
where $\textit{\textbf{f}}^t_i$ is the intermediate feature extracted from the input image, and $\textit{\textbf{f}}^t_o=\mathcal{F}_A^t(\textit{\textbf{f}}^t_i)$ is the reconstructed feature representation by the autoencoder of task $t$. Since the autoencoder $\mathcal{F}_A^t$ is individually learned on the data of task $t$, the resulting reconstruction score $d^t$ reflects the likelihood that an input image pertains to the task, with a lower score suggesting a higher probability.

Moreover, to preserve CLIP's zero-shot transfer ability during continual learning, we include an additional autoencoder, $\mathcal{F}^0_A$, trained on a reference dataset to identify out-of-distribution data. Upon completion of the learning process, DDAS computes a set of distribution scores $\{d^t\}_{t=1}^T$ for each input image. Should all scores surpass a specific threshold, $Thres$, the system classifies the input as ``unseen data" and redirects it to the frozen CLIP for zero-shot transfer. Conversely, inputs below this threshold are routed to the corresponding router with the lowest distribution score, ensuring efficient and accurate task identification.

%% file: sec/4_experiments.tex
\section{Experiments}
\subsection{Experimental Setting}
\label{ablation study}
\textbf{Datasets.}
We evaluate our method across two tasks: Multi-domain TIL (MTIL) and CIL. For MTIL, we follow the two-order training protocol proposed in \cite{zheng2023preventing}. For CIL, we follow~\cite{douillard2022dytox} to conduct experiments on CIFAR100 \cite{douillard2022dytox} and TinyImageNet \cite{yan2021dynamically}. The 100 classes of CIFAR100 are divided into $\{10, 20, 50\}$ subsets, and the 100 classes from TinyImageNet are divided into $\{ 5, 10, 20\}$ subsets to evaluate class distribution adaptability.

\noindent\textbf{Metrics.}
To evaluate our method on the MTIL, we utilize metrics proposed by \cite{zheng2023preventing}, namely ``Transfer'', ``Average'', and ``Last''. The ``Transfer'' metric assesses the model's zero-shot transfer capability on unseen data. ``Last'' evaluates the model's memorization ability on historical knowledge. ``Average'' is a composite metric measuring the mean performance across ``Transfer'' and ``Last''. In CIL, following \cite{douillard2022dytox}, we calculate the average accuracy over all subsets (``Average'') and specifically for the last subset (``Last'').

\noindent\textbf{Implementation Details.}
As in \cite{zheng2023preventing}, we use the CLIP model with ViT-B/16~\cite{dosovitskiy2020image} as our backbone for all the experiments. 
We adopt LoRA~\cite{hu2021lora} as experts and set the total number $N_E=22$. The router is a single MLP that mixes the experts with $top$-$2$ gating scores. In DDAS, the reference data is TinyImageNet~\cite{yan2021dynamically}, and the threshold is $0.065$ and $0.06$ for full-shot and few-shot. The autoencoder is built upon a pretrained AlexNet~\cite{krizhevsky2012imagenet} with MLP and a non-linear layer. We use AdamW~\cite{loshchilov2017decoupled} optimizer and a label smoothing~\cite{muller2019does} technique for a better result.
For TIL, we train 1k iterations on full-shot and 500 iterations on few-shot for each task. For DDAS, we train 1k and 300 iterations for reference datasets and incremental tasks, respectively. Except for the reference dataset, the MoE-Adapters and DDAS are jointly trained during continual learning. %

\subsection{Comparison with State-of-the-art Methods}
\label{comparison}
\textbf{Multi-domain Task Incremental Learning.} 
Table~\ref{tab:compareAvg_Last_Transfer} showcases a comparison between our proposed method and alternative approaches in the MTIL task. Our approach utilizes the predefined Order-\uppercase\expandafter{\romannumeral1} from~\cite{zheng2023preventing}, where datasets are trained and tested sequentially in a left-to-right order, as displayed in Table~\ref{tab:compareAvg_Last_Transfer}. Additional results for Order-\uppercase\expandafter{\romannumeral2} are provided in the supplementary material. 
The uppermost section of Table~\ref{tab:compareAvg_Last_Transfer} displays the outcomes of applying CLIP independently on each task through zero-shot inference, full parameter fine-tuning, and parameter-efficient fine-tuning. The ``zero-shot" represents the optimal results of CLIP's zero-shot transfer on each task, while the other two rows indicate the highest possible outcomes achieved by fine-tuning CLIP in each respective task.
Our proposed method, labeled as ``Ours", outperforms the second-best approach on most tasks, resulting in an overall improvement of by 0.8\%, 1.3\%, and 1.4\% in terms of ``Transfer'', ``Average'' and ``Last'', respectively. Additionally, by increasing the training iterations from 1k to 3k, our method (labelled as ``Ours$\dagger$'') achieves a further improvement of 1.8\% and 3.8\% on ``Average'' and ``Last'' while dropping 0.7\% on ``Transfer'' in comparison to ZSCL~\cite{zheng2023preventing}. Furthermore, our model achieves less degradation than ZSCL when compared with the upper bound methods, demonstrating favorable performance in anti-forgetting and zero-shot transfer.

\begin{table}[t]
        \centering
        \resizebox{0.95\linewidth}{!}{
	
	\begin{tabular}{lcc|cc|cc}
            \toprule 
            \multirow{2}{*}{Method} & \multicolumn{2}{c}{10 step} & \multicolumn{2}{c}{20 step} & \multicolumn{2}{c}{50 step} \\
            \cline{2-7} 
            \noalign{\smallskip}
			 & 	Avg. & Last &Avg. & Last &Avg. &Last  \\
		\midrule
            UCIR \cite{hou2019learning} & 58.66 & 43.39& 58.17 & 40.63 & 56.86 & 37.09 \\
            Bic\cite{wu2019large}&68.80 &53.54 &66.48&47.02 & 62.09 & 41.04\\
		PODNet\cite{douillard2020podnet} &58.03&41.05&53.97&35.02&51.19&32.99\\
            DER \cite{yan2021dynamically} & 74.64 & 64.35 & 73.98 & 62.55 & 72.05 & 59.76\\
            DyTox+\cite{douillard2022dytox} & 74.10 & 62.34 & 71.62 & 57.43 & 68.90 & 51.09 \\
            DNE \cite{hu2023dense} & 74.86& 70.04& - & - & -& - \\
            \midrule
            CLIP Zero-shot &74.47  & 65.92 & 75.20 & 65.74  & 75.67 & 65.94 \\
            Fine-tune & 65.46& 53.23& 59.69& 43.13&39.23 &18.89\\
            LwF \cite{li2017learning}&65.86 &48.04 &60.64 &40.56 &47.69 &32.90\\
            iCaRL  \cite{rebuffi2017icarl} & 79.35 &70.97&73.32 &64.55 & 71.28&59.07\\
            LwF-VR \cite{ding2022don}&78.81 &70.75 &74.54 &63.54 &71.02 & 59.45\\
            ZSCL \cite{zheng2023preventing} &\underline{82.15} &\underline{73.65} &\underline{80.39} &\underline{69.58} &\underline{79.92} &\underline{67.36}\\
        \rowcolor{Thistle!20}
		{Ours} & \textbf{85.21} &\textbf{77.52}& \textbf{83.72}&\textbf{76.20} & \textbf{83.60} & \textbf{75.24}  \\
		\bottomrule
	\end{tabular}}
    \vspace{-5pt}
	\caption{Comparison of different methods on CIFAR100 in class-incremental setting. We label the best and second-best methods with \textbf{bold} and \underline{underline} styles.}
	\label{tab:CIL1}
 \vspace{-9pt}
\end{table}
\begin{table}[t]
	\centering
         \resizebox{0.95\linewidth}{!}{
	\begin{tabular}{lcc|cc|cc}
            \toprule \multirow{2}{*}{Method} & \multicolumn{2}{c}{5 step} & \multicolumn{2}{c}{10 step} & \multicolumn{2}{c}{20 step} \\
            \cline{2-7} 
            \noalign{\smallskip}
		 & 	Avg. & Last &Avg. & Last & Avg. &Last  \\
		\midrule
            EWC \cite{kirkpatrick2017overcoming} & 19.01 & 6.00&15.82 &3.79 & 12.35&4.73  \\
            EEIL \cite{castro2018end} &47.17 & 35.12&45.03 &34.64 & 40.41&29.72\\
            UCIR \cite{hou2019learning}&50.30 & 39.42& 48.58& 37.29& 42.84& 30.85  \\
		  MUC \cite{liu2020more} &32.23&19.20 & 26.67& 15.33& 21.89&10.32  \\
            PASS \cite{zhu2021prototype} &49.54 & 41.64&47.19& 39.27& 42.01& 32.93 \\
            DyTox \cite{douillard2022dytox} & 55.58& 47.23&52.26 & 42.79& 46.18&36.21   \\
            
            \midrule
            CLIP Zero-shot & 69.62& 65.30& 69.55& 65.59& 69.49&65.30  \\
            Fine-tune &61.54 &46.66 &57.05 & 41.54& 54.62 & 44.55  \\
            LwF \cite{li2017learning}&60.97 &48.77 &57.60 &44.00 &54.79 &42.26  \\
            iCaRL  \cite{rebuffi2017icarl} & 77.02& 70.39& 73.48&65.97 &69.65 &64.68  \\
            LwF-VR \cite{ding2022don}& 77.56&70.89 &74.12 &67.05 & 69.94&63.89  \\
            ZSCL \cite{zheng2023preventing}& \underline{80.27}& \underline{73.57}&\underline{78.61} & \underline{71.62}& \underline{77.18}&\underline{68.30}  \\
        \rowcolor{Thistle!20}
		Ours& \textbf{81.12} & \textbf{76.81}& \textbf{80.23}& \textbf{76.35}& \textbf{79.96} & \textbf{75.77}   \\
		\bottomrule
	\end{tabular}}
 \vspace{-4pt}
	\caption{Comparison of different methods on TinyImageNet dataset in class-incremental settings with 100 base classes. We label the best and second methods with \textbf{bold} and \underline{underline} styles.}
	\label{tab:CIL2}
 \vspace{-11pt}
\end{table}

\noindent\textbf{Few-shot Multi-Domain Task Incremental Learning.} 
we also ran experiments on few-shot MTIL, limiting the CLIP model to access only a few samples per task. In a 5-shot setting, the comparison results are shown in Table~\ref{tab:comparefs} using the same metrics as Table~\ref{tab:compareAvg_Last_Transfer}. Our method outperforms most state-of-the-art approaches on most datasets, surpassing the second-best method by 3.6\%, 7.0\%, and 4.2\% in terms of ``Transfer," ``Average," and ``Last". These results demonstrate the effectiveness of the proposed incremental MoE-Adapters in addressing the forgetting issue in long-term continual learning, even with limited samples. Furthermore, our proposed DDAS effectively learns data distribution discrimination with fewer training samples.

\noindent\textbf{Class Incremental Learning.}
We conduct experiments on class incremental learning to verify our method's performance on single-domain CL. Unlike MTIL, the task id of the input image is unknown in CL. To this end, our MoE-Adapters use only one router with two experts to adapt to all the subsets. The comparison results between our method and state-of-the-art approaches on CIFAR100 and TinyImageNet are shown in Table~\ref{tab:CIL1} and~\ref{tab:CIL2}, respectively. As we can see, the proposed method consistently outperforms the other competitors, including dynamic expansion and CLIP-based approaches, demonstrating the effectiveness and scalability of our MoE-Adapters in addressing single-domain CL. 

\begin{table}[t]
	\centering
         \resizebox{0.96\linewidth}{!}{
	\begin{tabular}{c >{\centering\arraybackslash}p{2.4cm} >{\centering\arraybackslash}p{2.1cm}>{\centering\arraybackslash}p{1.8cm}}
		\toprule		Method & Train Params $\downarrow$ &GPU $\downarrow$ & Times $\downarrow$ \\
		\midrule
            \textcolor{Black}{LWF}~\cite{li2017learning}&\textcolor{Black}{149.6M} & \textcolor{Black}{32172MiB} & \textcolor{Black}{1.54s/it}\\
            \textcolor{Black}{LWF-VR}~\cite{ding2022don} & \textcolor{Black}{149.6M} & \textcolor{Black}{32236MiB} &\textcolor{Black}{1.51s/it}\\
            \rowcolor{gray!20}
		ZSCL~\cite{zheng2023preventing} &149.6M & 26290MiB & 3.94s/it \\
		MoE-Adapters &\ \ 51.1M & 19898MiB & 1.37s/it \\
		DDAS  & \ \ 8.7M & \ \ 2461MiB & 0.21s/it \\
            \rowcolor{Thistle!20}		
            Ours&	\ \ 59.8M&	22358MiB	& 1.58s/it \\
            \textbf{$\Delta$} &	\textbf{\textcolor{Maroon}{-60.03\%}}&	\textbf{\textcolor{Maroon}{-14.95\%}}	& \textbf{\textcolor{Maroon}{-59.90\%}} \\
		\bottomrule
	\end{tabular}}
	\caption{\textcolor{Black}{Comparison of computational cost during training between our method and others in terms of training parameters, GPU burdens and training times of each iteration. And the \textbf{$\Delta$} is the improvement relative to the SOTA ZSCL~\cite{zheng2023preventing}.}}
	\label{tab:cost}
 \vspace{-10pt}
\end{table}

\noindent\textbf{Computational Cost.}
The experiments above have demonstrated the promising performance of our method in both MTIL and CIL. We further compare the computational cost of our method with \textcolor{Black}{others} to prove its parameter and time efficiency during training. Table~\ref{tab:cost} shows that our method is superior to \textcolor{Black}{the SOTA method ZSCL}, with a reduction of approximately 60\%, 15\%, and 60\% in terms of training parameter (M), GPU burdens (MiB), and iteration time, respectively. Additionally, we analyze the efficiency of our two proposed components, MoE-Adapters and DDAS, demonstrating that they effectively enhance continual learning for CLIP while reducing significant computation burdens during training.


\subsection{Ablation Study}
\label{ablation}
In this section, we mainly analyze the efficacy of the proposed incremental MoE-Adapters and DDAS. All the experiments are conducted in MTIL setting. More analysis can be found in the supplementary material.   

\noindent\textbf{Analysis of MoE-Adapters.}
We conduct detailed ablation studies of different settings on MoE-Adapters, as shown in Table \ref{tab:ablation1}. Compared with the zero-shot CLIP and the fine-tuned version with one adapter, our MoE-Adapters effectively mitigate the ``catastrophic forgetting'' issue and retain the zero-shot transfer ability on the unseen date. In the proposed MoE-Adapters, we use $T$ task-specific routers to adaptively activate the $Topk$ experts from the predefined expert pool. Table~\ref{tab:ablation1} illustrates several different experts and routers combinations. As we can see, compared with using more experts, the task-specific routers contribute more to improve anti-forgetting and zero-shot transfer abilities. In the training of MoE, we propose an incremental activate-freeze strategy, enabling the collaboration of previously learned experts and inactivated ones for more accurate prediction. The comparison between ``Ours'' and ``+22E/11R w/o F'' demonstrates the effectiveness of this strategy.

\begin{table}[t]
	\centering
        \resizebox{0.95\linewidth}{!}{
	\begin{tabular}{lcc|cc|cc}
		\toprule		
            Method &  Transfer & $\Delta$ &Avg.&  $\Delta$ &  Last &$\Delta$ \\
		\midrule
            CLIP Zero-shot &  69.4 & \textcolor{Maroon}{+0.5} & 65.3 &  \textcolor{NavyBlue}{-11.4} &  65.3 &  \textcolor{NavyBlue}{-19.7}\\
		{\hspace{1em}}+Adapter &45.0 & \textcolor{NavyBlue}{-23.9}  & 57.0 & \textcolor{NavyBlue}{-19.7} &  71.5 & \textcolor{NavyBlue}{-13.5}  \\
		{\hspace{1em}}+2E/1R &45.1  &\textcolor{NavyBlue}{-23.8} &56.3 &\textcolor{NavyBlue}{-20.4}  & 71.1& \textcolor{NavyBlue}{-13.9}\\
            {\hspace{1em}}+2E/11R & 68.1 & \textcolor{NavyBlue}{-0.8} & 72.6&  \textcolor{NavyBlue}{-4.1}  & 77.9&  \textcolor{NavyBlue}{-7.1}\\
            {\hspace{1em}}+22E/1R & 44.1 & \textcolor{NavyBlue}{-24.8}&56.0 & \textcolor{NavyBlue}{-20.7} & 66.2& \textcolor{NavyBlue}{-18.8}\\
            {\hspace{1em}}+22E/11R w/o F & 68.6  &\textcolor{NavyBlue}{-0.3} & 75.1 &  \textcolor{NavyBlue}{-1.6} & 82.0 & \textcolor{NavyBlue}{-3.0}\\
            \rowcolor{Thistle!20}
            {\hspace{1em}}Ours & 68.9& \ 0.0 & 76.7&  \ 0.0 &  85.0 & \ 0.0 \\
		\bottomrule
	\end{tabular}}
 \vspace{-5pt}
	\caption{Ablation studies on incremental MoE-Adapters. ``mE/nR'' indicates MoE with m experts and n routers, respectively. ``F'' represents the incremental activate-freeze strategy. }
  \vspace{-5pt}
	\label{tab:ablation1}
\end{table}

\begin{figure}[t]
	\centering
	\includegraphics[width=0.95\linewidth]{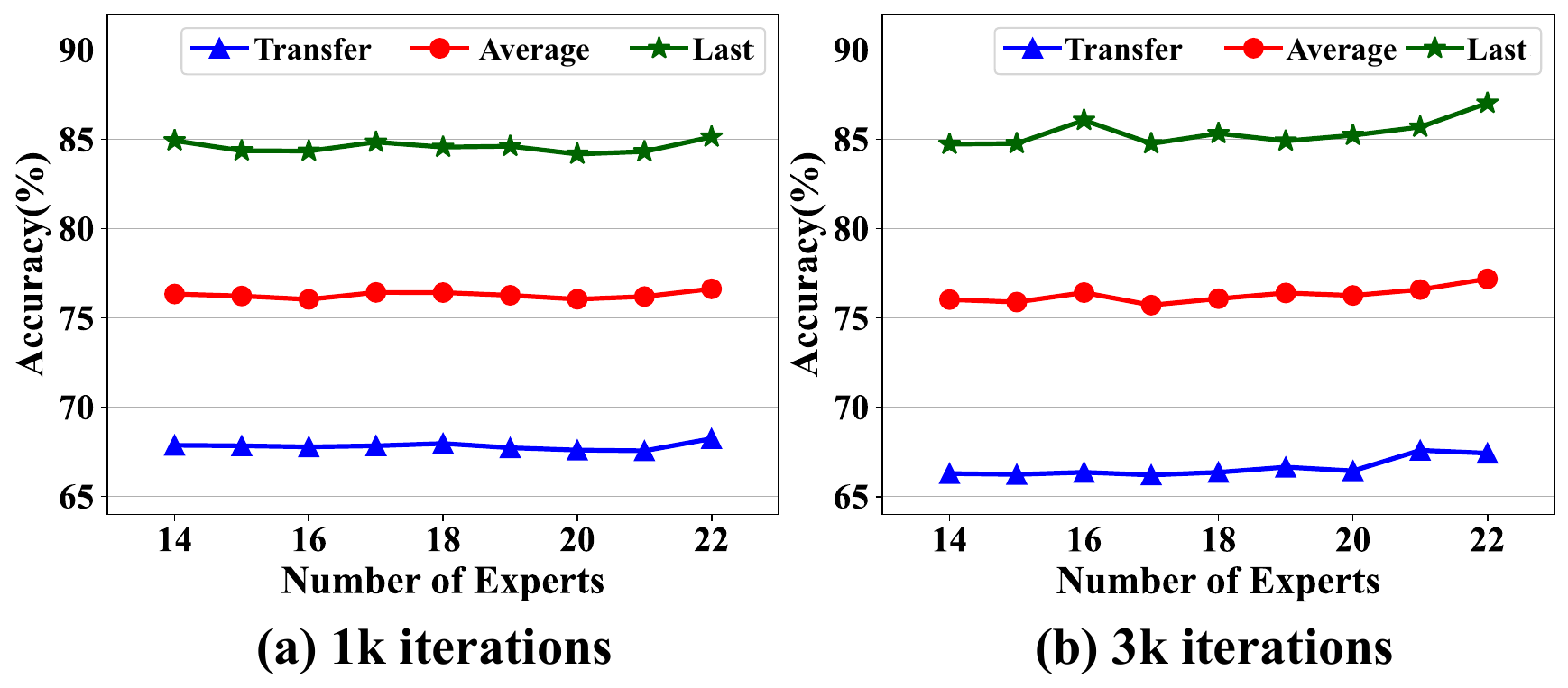}
 \vspace{-5pt}
	\caption{Analysis of expert's number in different training iterations. 
    The results can be referred to ``Ours'' and ``Ours$\dagger$'' in Table~\ref{tab:compareAvg_Last_Transfer}.
    }
	\label{fig:ablation}
  \vspace{-12pt}
\end{figure}

\noindent\textbf{Analysis of Expert Number.}
Figure~\ref{fig:ablation} presents the ablation study on the number of experts used in MoE-Adapters. The experiments are conducted on the models trained by 1k~/~3k iterations with $T$ task-specific routers. The smoothness of the curves in the figure indicates our method's robustness for changing the number of experts. We can observe that the three metrics remain relatively stable as the number of experts changes. 
As shown in Figure~\ref{fig:ablation} (a) and (b), it is more stable in the 1k iteration setting than in 3k iterations. This phenomenon arises due to the escalating number of iterations, leading to an increase in the overall frequency of expert selection. When the number of experts is small, our activate-freeze strategy may activate more untapped experts and train them a few times, leading to the router mistakenly activating incompletely trained experts during the inference phase. The fluctuation is within an acceptable range and does not significantly affect the final performance.


\begin{figure}[t]
	\centering
	\includegraphics[width=1\linewidth]{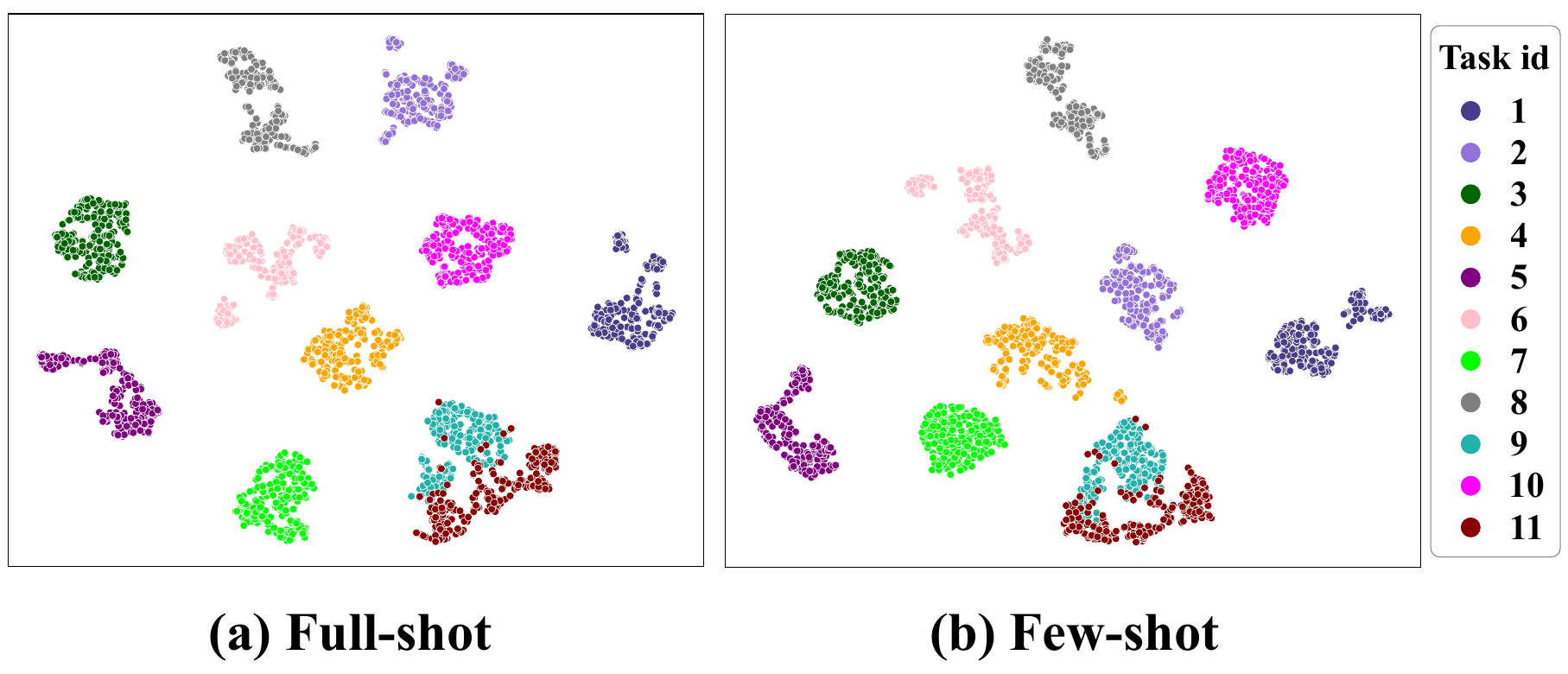}
 \vspace{-20pt}
	\caption{t-SNE on DDAS's output of each task on full-shot and few-shot MTIL. The corresponding task names from $id=1-11$ are matches with the datasets listed from left to right in Table~\ref{tab:compareAvg_Last_Transfer}. }
	\label{fig:tsne}
  \vspace{-12pt}
\end{figure}

\noindent\textbf{Analysis of DDAS.}
We use DDAS to automatically distinguish the input images from seen or unseen data by learning the variations in data distribution with task-specific autoencoders. To verify the effectiveness of DDAS, we analyze the distribution discrimination in feature space, whose results are illustrated in Figure~\ref{fig:tsne}. We employ the reconstructed features $\textit{\textbf{f}}^t_o$ and plot the figure when the continual learning is finished. As we can see, the proposed DDAS is effective at learning the discriminative distribution of each learned task in full-shot and few-shot MTIL. 
As shown in Figure~\ref{fig:tsne}, the feature distributions of some samples from Task 9 overlap with that of Task 11. It is because these samples are misclassified by DDAS as out-of-distribution data and perform feature extraction with reference autoencoder. Although the inevitable misclassifications occur, our method still outperforms the state-of-the-art approaches in various metrics.

%% file: sec/5_conclusion.tex
\section{Discussion}
We propose a parameter-efficient training framework to boost the continual learning of vision language models. We employ MoE-Adapters to help the CLIP model to adapt efficiently and generalize well on all tasks. 
Moreover, we introduce a Distribution Discriminative Auto-Selector (DDAS) to assign inference data automatically to either MoE-Adapters or the frozen CLIP.  
Extensive experimental results in various settings demonstrate the superiority of our method over previous arts in terms of classification accuracy and training efficiency. 

One limitation of our framework is that the proposed DDAS requires a predefined threshold to determine downstream branches for all tasks. With the growth of task numbers, adapting all tasks with a single threshold would bring errors. Besides, incorporating the learned knowledge to improve the zero-shot transfer ability of the original CLIP is a future research direction.

\noindent{\textbf{Acknowledgement.} This work was supported by National Key R\&D Program of China under Grant 2021YFA0715202, National Natural Science Foundation of China under Grant 62206039 and 62293544, the Fundamental Research Funds for the Central Universities, and CCF-Baidu Open Fund (no. CCF-Baidu202314).}

%% file: X_suppl.tex
\clearpage
\setcounter{page}{1}
\maketitlesupplementary


\section*{A. More Implementation Details}
\label{sec:More Implementation Details}
We set batch size as $64$ for the Multi-domain Task Incremental Learning (MTIL) benchmark and $128$ for the Class Incremental Learning (CIL) benchmark. The learning rates are searched among $[10^{-3}, 10^{-4}]$. Label smoothing can substitute the regularization of weight decay and achieve better performance. The label smoothing strength is searched between $\{0.1, 0.2\}$. For CIL, we set weight decay as $0$ and label smoothing as $0.0$.
\section*{\textcolor{black}{B. Impact of dataset size on expert number}}
\textcolor{black}{The additional ablation experiments are conducted to explore the optimal number of experts for different task size, and the results are shown in Table~\ref{tab:size_number}. The table showcases the impact of dataset size on the optimal number of experts $N_E$ in full-shot setting. We notice that, in general, more tasks require more experts, while simply applying more experts does not always improve accuracy.}

\section*{\textcolor{black}{C. Analysis on the Threshold and Different Loss in DDAS}}
\label{sec:Additional Analysis of different thresholds}
\begin{figure}[h]
	\centering
	\includegraphics[width=0.95\linewidth]{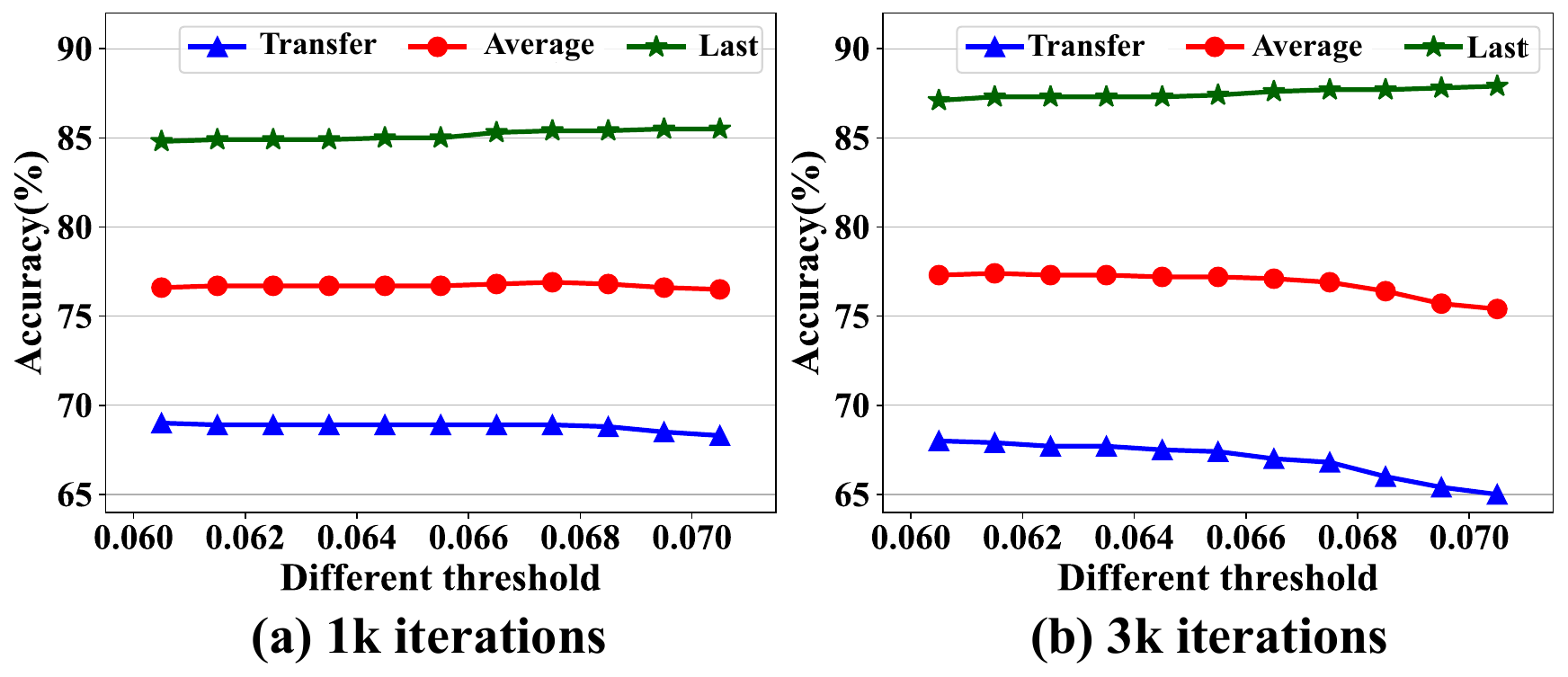}
	
	\caption{The ablation study of different thresholds in DDAS, and the thresholds are searched within the range of $[0.06, 0.07]$. 
}
	\label{fig:suppl}
\end{figure}

To further analyze the impact of different thresholds ($Thres$) in the Distribution Discriminative Auto-Selector (DDAS), we perform ablation experiments with different thresholds in the methods (``Ours'' and ``Ours$\dagger$''), which are shown in Figure~\ref{fig:suppl}. The thresholds are searched within the range of $[0.06, 0.07]$.
The results show that the performance fluctuation of our method is relatively stable within a certain threshold range. Compared with the method ``Ours$\dagger$'', the method ``Ours'' demonstrates more consistent performance as the threshold changes.

\textcolor{black}{In addition, we conduct ablation experiments on various loss functions for the autoencoder of DDAS, and the results are shown in Table~\ref{tab:different loss}. It can be seen that our method achieve the best performance when utilizing the Mean Squared Error (MSE) loss.}

\section*{D. More Comparison Results on MTIL}
\label{sec:More comparison results on MTIL}
The complete result of the MTIL benchmark with $T$ datasets is a matrix of $T$ × $T$, where $T$ is the number of incremental tasks. 
In Table~\ref{tab:suppl1} and~\ref{tab:suppl2}, we present the complete matrices of both ``Ours'' (trained in 1k iterations) and ``Ours$\dagger$'' (trained in 3k iterations) for the MTIL benchmark. 
In addition, Table~\ref{tab:suppl3} and~\ref{tab:suppl4-fewshot} show the results of the full-shot and few-shot MTIL benchmarks in Order-II. The Order-II sequence includes: StanfordCars, Food, MNIST, OxfordPet, Flowers, SUN397, Aircraft, Caltech101, DTD, EuroSAT, CIFAR100. As we can see, the proposed method performs favorably against state-of-the-art approaches in terms of three metrics in both settings. Notably, the zero-shot transfer ability of the proposed method closely reaches the upper bound of the pretrained CLIP.

\section*{E. Effectiveness of Router Selection in MoE-Adapters}
\label{sec: More visualization on MoE-Adapters}
\begin{figure*}[t]
	\centering
	\includegraphics[width=0.95\linewidth]{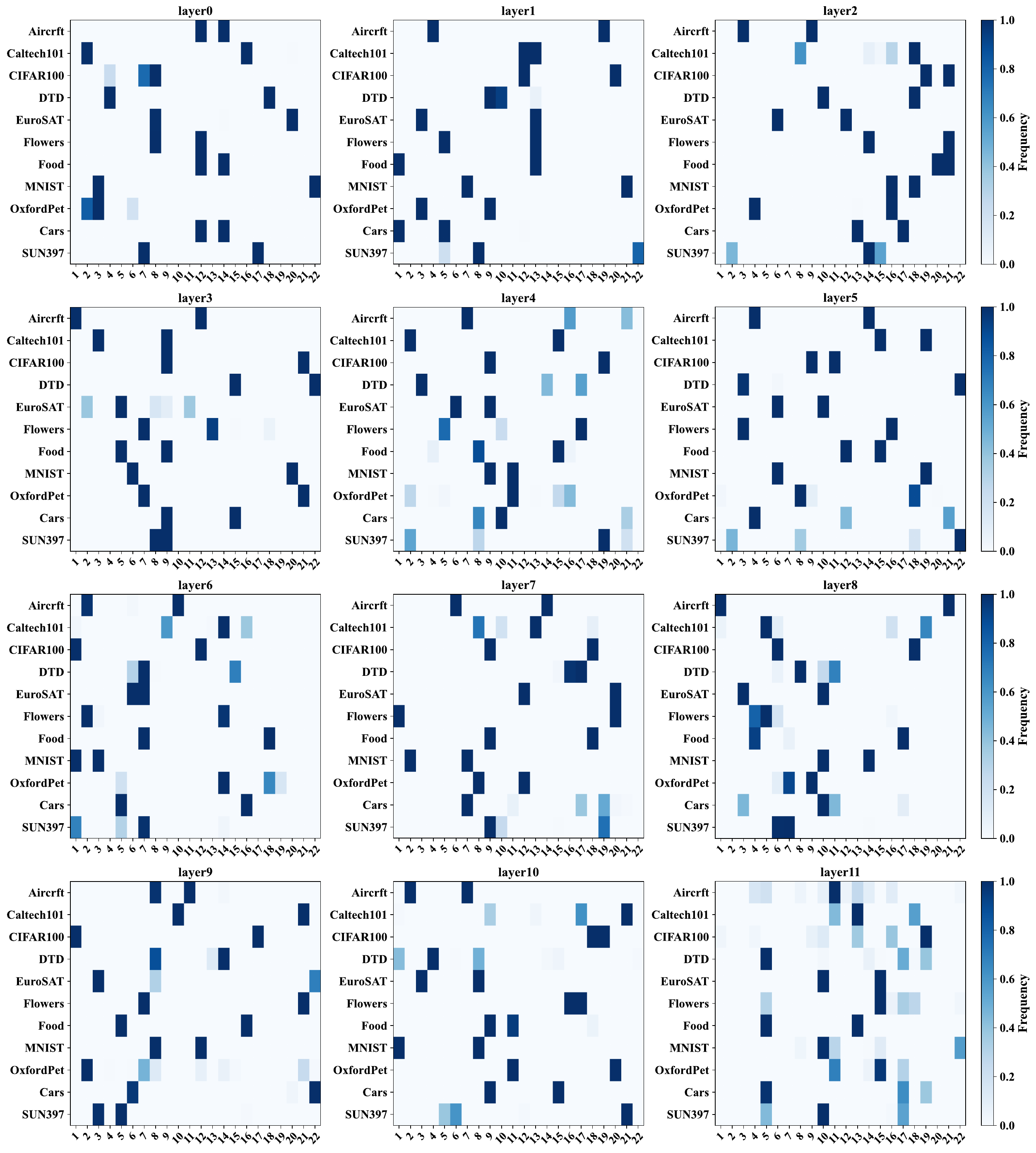}
	
	\caption{Visualization of the frequency that experts are selected for each task in task incremental learning. The activation frequencies of MoE-Adapters' experts are recorded in all transformer blocks of the visual encoder, with 22 experts and $Top$-$K$ as 2. The $y$-axis represents incremental tasks and the $x$-axis represents the experts.}
	\label{fig:suppl2}
\end{figure*}

We visualize the frequency that MoE-Adapters' experts are selected for each incremental task, as shown in Figure~\ref{fig:suppl2}. 
As we can see, the activation frequencies of experts are recorded in all visual transformer blocks of CLIP, with 22 experts for each block and $Top$-$k$ as 2. 
The visualization demonstrates the sparsity of the experts activated by our router selection and the cooperation between special experts and shared experts.

\vspace{-3pt}
\begin{table}[h]
    \centering
    \resizebox{0.99\linewidth}{!}{
    \begin{tabular}{>{\centering\arraybackslash}p{0.6cm}>{\centering\arraybackslash}p{0.6cm}>{\centering\arraybackslash}p{0.6cm}>{\centering\arraybackslash}p{0.6cm}>{\centering\arraybackslash}p{0.6cm}>{\centering\arraybackslash}p{0.6cm}>{\centering\arraybackslash}p{0.6cm}>{\centering\arraybackslash}p{0.6cm}>{\centering\arraybackslash}p{0.6cm}>{\centering\arraybackslash}p{0.6cm}}
    \specialrule{\lightrulewidth}{0em}{0em}
         \multirow{2}{*}{$N_E$} & \multicolumn{3}{c}{4-task} &  \multicolumn{3}{c}{8-task}&  \multicolumn{3}{c}{11-task}  \\
            \cline{2-4} \cline{5-7} \cline{8-10}

			 & 	Trans. & Avg. & Last & Trans. & Avg. & Last & Trans. & Avg. & Last \\
    \specialrule{\lightrulewidth}{0em}{0em}
          2&65.8&60.7&59.0&  65.9&63.2&63.2&67.3&64.1&61.5 \\  
          4&64.9&67.5&77.1& 65.0&71.2&77.9&66.5&71.1&74.1\\ 
          8&\cellcolor{Thistle!20}65.1&\cellcolor{Thistle!20}68.3&\cellcolor{Thistle!20}78.3& 65.4&73.7&84.9&67.4&75.7&82.4\\  
          16&65.3&67.7&77.9& 65.5 &73.9&84.9&\cellcolor{Thistle!20}68.0&\cellcolor{Thistle!20}76.4&\cellcolor{Thistle!20}84.6\\
          20 &65.5&67.4&77.0&\cellcolor{Thistle!20}66.6&\cellcolor{Thistle!20}74.6&\cellcolor{Thistle!20}85.8&67.6&76.0&84.2 \\
    \specialrule{\lightrulewidth}{0em}{0em}
    \end{tabular}}
    \caption{\textcolor{black}{Ablation study on the number of experts across different size of dataset.}}
    \label{tab:size_number}
\end{table}
\vspace{-13pt}
\begin{table}[h]
    \centering
    \resizebox{0.90\linewidth}{!}{
    \begin{tabular}{c>{\centering}p{0.85cm}>{\centering\arraybackslash}p{0.85cm}>{\centering\arraybackslash}p{0.85cm}>{\centering\arraybackslash}p{0.85cm}>{\centering\arraybackslash}p{0.85cm}>{\centering\arraybackslash}p{0.85cm}}
    \toprule
    \multirow{2}{*}{Method} & \multicolumn{3}{c}{full-shot} & \multicolumn{3}{c}{5-shot}  \\
            \cline{2-7}
            \vspace{-1pt}
			 &Trans. & Avg. & Last & Trans. & Avg. & Last  \\
    \midrule
          \rowcolor{gray!20}
          ZSCL[78]&68.1&75.4&83.6&65.3&66.7&67.4\\
          MAE  & 68.4 &73.8 &  77.8&68.5 & 70.9&70.5 \\
          Smooth L1  & 68.3 & 76.9 & 84.9 &69.0 & 72.9&72.6 \\
          \rowcolor{Thistle!20}
          MSE  & 68.9 & 76.7 &85.0 & 68.9&76.3 & 76.1\\
          \specialrule{\lightrulewidth}{0em}{0em}
    \end{tabular}}
    \caption{\textcolor{black}{Ablation study of different loss in DDAS. }}
    \label{tab:different loss}
        
\end{table}

\vspace{-11pt}

\vspace{-10pt}

\begin{table*}[h]
    \centering
    \resizebox{0.90\linewidth}{!}{
    \begin{tabular}{l>{\centering\arraybackslash}p{1cm} >{\centering\arraybackslash}p{1cm} >{\centering\arraybackslash}p{1cm}>{\centering\arraybackslash}p{1cm} >{\centering\arraybackslash}p{1cm} >{\centering\arraybackslash}p{1cm}>{\centering\arraybackslash}p{1cm} >{\centering\arraybackslash}p{1cm} >{\centering\arraybackslash}p{1cm}>{\centering\arraybackslash}p{1cm} >{\centering\arraybackslash}p{1cm} >{\centering\arraybackslash}p{1cm}}
        \toprule
             & \rotatebox{90}{Aircraft~\cite{maji2013fine}} & \rotatebox{90}{Caltech101~\cite{fei2004learning}} & \rotatebox{90}{CIFAR100~\cite{krizhevsky2009learning}} & \rotatebox{90}{DTD~\cite{cimpoi2014describing}}& \rotatebox{90}{EuroSAT~\cite{helber2019eurosat}} & \rotatebox{90}{Flowers~\cite{nilsback2008automated}} & \rotatebox{90}{Food~\cite{bossard2014food}} & {\rotatebox{90}{MNIST~\cite{deng2012mnist}}} & {\rotatebox{90}{OxfordPet~\cite{parkhi2012cats}}} & {\rotatebox{90}{Cars~\cite{krause20133d}}} & {\rotatebox{90}{SUN397~\cite{xiao2010sun}}} & \\
            \midrule
            Transfer & & 87.9 & 68.2 & 44.4 & 50.0 & 70.7 & 88.7 & 59.7 & 89.1 & 64.5 & 65.5 & \colorbox{Orchid}{68.9}\\
            \midrule
            Aircraft & \colorbox{YellowGreen}{51.5} & 87.9 & 68.2 & 45.1 & 54.6 &71.3 & 88.8&59.5 &89.1 &64.5 &65.3 & \\
            Caltech101 &51.0 &\colorbox{YellowGreen}{92.3} & 68.2&44.0 &54.6 &70.2&88.8&59.5 & 89.1 &64.5 &65.5 & \\
            CIFAR100 & 50.0&91.5&\colorbox{YellowGreen}{86.7}&44.2&44.4&70.8&88.8&59.8&89.1&64.5&65.5&\\
            DTD &50.4 &92.0&86.5&\colorbox{YellowGreen}{78.6}&45.9&70.7&88.8&59.8&89.1&64.5&65.5 &\\
            EuroSAT &50.4&91.8&86.5&78.3&\colorbox{YellowGreen}{96.1}&70.4&88.8&59.8&89.1&64.5&65.6&\\
            Flowers &50.3&92.3&86.3&79.1&95.7&\colorbox{YellowGreen}{95.9}&88.7&59.8&89.1&64.5&65.6& \\
            Food &49.7&93.0&86.4&78.9&95.3&95.8&\colorbox{YellowGreen}{89.5}&59.8&89.1&64.5&65.7&\\
            MNIST &49.7&92.7&86.3&79.0&95.5&95.6&89.5&\colorbox{YellowGreen}{98.3}&89.1&64.5&65.6& \\
            OxfordPet&49.7&92.4&86.3&79.2&95.1&94.6&89.5&98.1&\colorbox{YellowGreen}{89.8}&64.5&65.5& \\
            Cars &49.4&92.4&86.2&78.9&94.8&94.7&89.5&98.2&89.7&\colorbox{YellowGreen}{81.9}&65.5&\\
            SUN397 &49.8&92.2&86.1&78.1&95.7&94.3&89.5&98.1&89.9&81.6&\colorbox{YellowGreen}{80.0}& \colorbox{Dandelion}{85.0} \\
            \midrule
            Average &50.2&91.9&83.1&69.4&78.9&84.0&89.1&73.7&89.3&67.7&66.9& \colorbox{Tan}{76.7}\\

        \bottomrule
    \end{tabular}}
    \caption{Accuracy (\%) of our method (Ours) on the MTIL benchmark with order-I. Each row represents the performance on every dataset of the model trained after the corresponding task. \colorbox{Orchid}{Transfer}, \colorbox{Tan}{Average}, and \colorbox{Dandelion}{Last} metrics are shown in color.}
    \label{tab:suppl1}
    
\end{table*}

\begin{table*}[h]
    \centering
    \resizebox{0.90\linewidth}{!}{
    \begin{tabular}{l>{\centering\arraybackslash}p{1cm} >{\centering\arraybackslash}p{1cm} >{\centering\arraybackslash}p{1cm}>{\centering\arraybackslash}p{1cm} >{\centering\arraybackslash}p{1cm} >{\centering\arraybackslash}p{1cm}>{\centering\arraybackslash}p{1cm} >{\centering\arraybackslash}p{1cm} >{\centering\arraybackslash}p{1cm}>{\centering\arraybackslash}p{1cm} >{\centering\arraybackslash}p{1cm} >{\centering\arraybackslash}p{1cm}}
        \toprule
             & \rotatebox{90}{Aircraft~\cite{maji2013fine}} & \rotatebox{90}{Caltech101~\cite{fei2004learning}} & \rotatebox{90}{CIFAR100~\cite{krizhevsky2009learning}} & \rotatebox{90}{DTD~\cite{cimpoi2014describing}}& \rotatebox{90}{EuroSAT~\cite{helber2019eurosat}} & \rotatebox{90}{Flowers~\cite{nilsback2008automated}} & \rotatebox{90}{Food~\cite{bossard2014food}} & {\rotatebox{90}{MNIST~\cite{deng2012mnist}}} & {\rotatebox{90}{OxfordPet~\cite{parkhi2012cats}}} & {\rotatebox{90}{Cars~\cite{krause20133d}}} & {\rotatebox{90}{SUN397~\cite{xiao2010sun}}} & \\
            \midrule
            Transfer & & 87.9 & 68.2 & 42.4 & 41.4 & 68.7 & 88.7 & 59.4& 89.1 & 64.5 & 64.0 & \colorbox{Orchid}{67.4}\\
            \midrule
            Aircraft & \colorbox{YellowGreen}{54.3} & 87.9 &68.2 & 45.1 & 54.6 & 71.3 & 88.8 & 59.5 & 89.1& 64.5 & 65.3 & \\
            Caltech101 &54.2 &\colorbox{YellowGreen}{92.0} & 68.2 & 40.7 & 54.6 & 67.7 & 88.7 & 59.5 & 89.1 & 64.5 & 63.5 & \\
            CIFAR100 & 54.3 & 91.6&\colorbox{YellowGreen}{88.8}&41.4&28.3&68.3&88.8&59.3&89.1&64.5&64.0\\
            DTD &54.3&91.7&88.8&\colorbox{YellowGreen}{80.0}&28.2&68.1&88.8&59.4&89.1&64.5&64.1\\
            EuroSAT &54.3&91.7&88.8&79.9&\colorbox{YellowGreen}{98.0}&68.1&88.8&59.4&89.1&64.5&64.0&\\
            Flowers &54.3&91.5&88.8&79.7&98.1&\colorbox{YellowGreen}{97.8}&88.5&59.4&89.1&64.5&63.9& \\
            Food &54.3&91.0&88.8&80.0&98.1&97.8&\colorbox{YellowGreen}{89.7}&59.3&89.1&64.5&63.8&\\
            MNIST &54.4&91.2&88.8&80.0&98.1&97.8&89.7&\colorbox{YellowGreen}{99.1}&89.1&64.5&63.8& \\
            OxfordPet&54.3&90.8&88.8&79.8&98.1&97.6&89.6&99.1&\colorbox{YellowGreen}{89.6}&64.5&63.4& \\
            Cars &54.2&90.8&88.8&80.2&98.1&97.5&89.6&99.1&89.5&\colorbox{YellowGreen}{89.2}&63.8&\\
            SUN397 &54.3&90.8&88.8&80.3&98.1&97.5&89.6&99.1&89.5&89.2&\colorbox{YellowGreen}{83.8}& \colorbox{Dandelion}{87.4} \\
            \midrule
            Average & 54.3 & 91.0  & 85.1 &69.7 & 77.5&84.5&89.1 &73.8&89.2&69.0&65.8 & \colorbox{Tan}{77.2}\\

        \bottomrule
    \end{tabular}}
    \caption{Accuracy (\%) of our method (Ours$\dagger$) on the MTIL benchmark with order-I. Each row represents the performance on every dataset of the model trained after the corresponding task. \colorbox{Orchid}{Transfer}, \colorbox{Tan}{Average}, and \colorbox{Dandelion}{Last} metrics are shown in color.}
    \label{tab:suppl2}
\end{table*}

\begin{table*}[h]
    \centering
    \resizebox{0.95\linewidth}{!}{
    \begin{tabular}{c>{\raggedright\arraybackslash}p{3cm} >{\centering\arraybackslash}p{1cm} >{\centering\arraybackslash}p{1cm}>{\centering\arraybackslash}p{1cm} >{\centering\arraybackslash}p{1cm} >{\centering\arraybackslash}p{1cm}>{\centering\arraybackslash}p{1cm} >{\centering\arraybackslash}p{1cm} >{\centering\arraybackslash}p{1cm}>{\centering\arraybackslash}p{1cm} >{\centering\arraybackslash}p{1cm} >{\centering\arraybackslash}p{1cm} >{\centering\arraybackslash}p{1.5cm}}
        \toprule
           & {\hspace{1em}} Method & \makecell[c]{\rotatebox{90}{Cars~\cite{krause20133d}}} & \makecell[c]{\rotatebox{90}{Food~\cite{bossard2014food}}} & \makecell[c]{\rotatebox{90}{MNIST~\cite{deng2012mnist}}} & \makecell[c]{\rotatebox{90}{OxfordPet~\cite{parkhi2012cats}}} & \makecell[c]{\rotatebox{90}{Flowers~\cite{nilsback2008automated}}} & \makecell[c]{\rotatebox{90}{SUN397~\cite{xiao2010sun}}} & \makecell[c]{\rotatebox{90}{Aircraft~\cite{maji2013fine}}} & \makecell[c]{\rotatebox{90}{Caltech101~\cite{fei2004learning}}} & \makecell[c]{\rotatebox{90}{DTD~\cite{cimpoi2014describing}}} & \makecell[c]{\rotatebox{90}{EuroSAT~\cite{helber2019eurosat}}} & \makecell[c]{\rotatebox{90}{CIFAR100~\cite{krizhevsky2009learning}}} & \makecell[c]{{\textit{Average}}} \\
  
        \midrule
        
            \multirow{3}{*}{\rotatebox{90}{CLIP}}& {\hspace{1em}}Zero-shot & 64.7 & 88.5 & 59.4 & 89.0 & 71.0 & 65.2 & 24.3 & 88.4 & 44.6 & 54.9 & 68.2 & 65.3  \\
            & {\hspace{1em}}Full Fine-tune & 89.6 & 92.7 & 99.6 & 94.7 & 97.5 & 81.8 & 62.0 & 95.1 & 79.5 & 98.9 & 89.6 & 89.2  \\
            & {\hspace{1em}}Fine-tune Adapter & 89.1 & 92.9 & 99.2 &94.1&97.0&82.7&56.8&92.6&79.0&98.4&89.4&88.3\\
            
            \midrule\midrule
            \multirow{8}{*}{\rotatebox{90}{\textbf{Transfer}}} &{\hspace{1em}}Continual-FT & & 85.9&\textbf{59.6}&57.9&40.0&46.7&11.1&70.0&30.5&26.6&37.7& 46.6 \\
            & {\hspace{1em}}LwF \cite{li2017learning} &&87.8&58.5&71.9&46.6&57.3&12.8&81.4&34.5&34.5&46.8& 53.2  \\
            & {\hspace{1em}}iCaRL \cite{rebuffi2017icarl} &  & 86.1 & 51.8 & 67.6 & 50.4 & 57.9 & 11.0 & 72.3 & 31.2 & 32.7 & 48.1 & 50.9\\
            & {\hspace{1em}}LwF-VR \cite{ding2022don} & &88.2&57.0&71.4& 50.0 & 58.0 & 13.0 & 82.0 & 34.4 & 29.3 & 47.6 &  53.1 \\
            & {\hspace{1em}}WiSE-FT \cite{wortsman2022robust} & &87.2 & 57.6 & 67.0 & 45.0 & 54.0 & 12.9 & 78.6 & 35.5 & 28.4 & 44.3 &   51.1\\
            & {\hspace{1em}}ZSCL \cite{zheng2023preventing} & & 88.3 & 57.5 & 84.7 & 68.1 & \underline{64.8} & \textbf{21.1} & \textbf{88.2} & \textbf{45.3} & \textbf{55.2} & \textbf{68.2} & 64.1  \\
            \rowcolor{Thistle!20}
            & {\hspace{1em}}Ours$\dagger$ & &\textbf{88.8} &\underline{59.5} &\textbf{89.1}  & \underline{69.4} & \textbf{65.3} & 15.0 & \underline{87.9} & \underline{43.9} & \underline{54.6} & \textbf{68.2 }& \underline{64.2}\textbf{\textcolor{Maroon}{(+0.1)}}  \\
            \rowcolor{Thistle!20}
            & {\hspace{1em}}Ours & &\textbf{88.8} &\underline{59.5} & \textbf{89.1} & \textbf{69.9} & 64.4 & \underline{18.1} & 86.9 & 43.7 & \underline{54.6} & \textbf{68.2} & \textbf{64.3\textcolor{Maroon}{(+0.2)}} \\
            \midrule 
            \multirow{8}{*}{\rotatebox{90}{\textbf{Average}}} &{\hspace{1em}}Continual-FT    & 42.1 & 70.5 & \textbf{92.2} & 80.1 & 54.5 & 59.1 & 19.8 & 78.3 & 41.0 & 38.1 & 42.3 & 56.2 \\
            & {\hspace{1em}}LwF \cite{li2017learning} & 49.0 & 77.0 & 92.1 & 85.9 & 66.5 & 67.2 & 20.9 & 84.7 & 44.6 & 45.5 & 50.5 & 62.2  \\
            & {\hspace{1em}}iCaRL \cite{rebuffi2017icarl} & 52.0 & 75.9 & 77.4 & 74.6 & 58.4 & 59.3 &  11.7& 79.6 & 42.1 & 43.2 & 51.7 & 56.9\\
            & {\hspace{1em}}LwF-VR \cite{ding2022don} & 44.9 & 75.8 & 91.8 & 85.3 & 63.5 & 67.6 & 16.9 & 84.9 & 44.0 & 40.6  & 51.3 &  60.6 \\
            & {\hspace{1em}}WiSE-FT \cite{wortsman2022robust} & 52.6 & 79.3 & 91.9 & 83.9  & 63.4 & 65.2 & 23.3 & 83.7  & 45.4 & 40.0 & 48.2 & 61.5\\
            & {\hspace{1em}}ZSCL \cite{zheng2023preventing} & 81.7& \textbf{91.3} & 91.1 & \underline{91.0} & 82.9 & \underline{72.5} & \textbf{33.6} & \underline{89.7} & \textbf{53.3} & \textbf{62.8} & \textbf{69.9} &   74.5 \\
            \rowcolor{Thistle!20}
            & {\hspace{1em}}Ours$\dagger$& \textbf{86.8} & 89.3 & \textbf{92.2} & 89.1 & \underline{86.0} & \textbf{73.0} & 30.8 & \textbf{90.0} & 53.1 & \underline{62.6} & \textbf{69.9} & \textbf{74.8}\textbf{\textcolor{Maroon}{(+0.3)}} \\
            \rowcolor{Thistle!20}
            & {\hspace{1em}}Ours & \underline{84.9} & \underline{89.9} & 89.3 & \textbf{91.4} & \textbf{86.2} & 72.2 & \underline{33.4} & 89.4 & \textbf{53.3} & 61.4 & \textbf{69.9} & \underline{74.7}\textbf{\textcolor{Maroon}{(+0.2)}}\\
            \midrule
            \multirow{8}{*}{\rotatebox{90}{\textbf{Last}}} &{\hspace{1em}}Continual-FT & 24.0& 67.3 &99.1 & 87.4 & 44.3 & 67.0 & 29.5 & 92.3 & 61.3 & 81.0 & \textbf{88.1} & 67.4 \\
            & {\hspace{1em}}LwF \cite{li2017learning} & 34.6 & 69.6 & 99.3 & 88.7 & 61.1 & 72.5 & 32.5 & 88.1 & 65.6 & 90.9 & 87.9 & 71.9 \\
            & {\hspace{1em}}iCaRL \cite{rebuffi2017icarl} & 46.0 & 81.5 & 91.3 & 82.8 & 66.5 & 72.2 & 16.3 & 91.6 & 68.1 & 83.2 & 87.8 &  71.6 \\
            & {\hspace{1em}}LwF-VR \cite{ding2022don}& 27.4 & 61.2 & \underline{99.4} & 86.3 & 60.6 & 70.7 & 23.4 & 88.0 & 61.3 & 84.3 & \textbf{88.1} & 68.2 \\
            & {\hspace{1em}}WiSE-FT \cite{wortsman2022robust} & 35.6 & 76.9 & \textbf{99.5} & 89.1 & 62.1 & 71.8 & 27.8 & 90.8 & 67.0 & 85.6 & 87.6 &  72.2 \\
            & {\hspace{1em}}ZSCL \cite{zheng2023preventing} & 78.2 & \textbf{91.1} & 97.6 & \textbf{92.5} & 87.4 & \textbf{78.2} & \underline{45.0} & 92.3 & 72.7 & \underline{96.2} & 86.3 &  83.4\\
            \rowcolor{Thistle!20}
            & {\hspace{1em}}Ours$\dagger$ & \underline{83.7} & \underline{89.0} & 99.2& 88.7 & \underline{92.9} & 75.9 & 42.6  & \underline{93.1} & \underline{76.6} & \textbf{98.6} &86.4& \textbf{84.2}\textbf{\textcolor{Maroon}{(+1.8)}} \\ 
            \rowcolor{Thistle!20}
            & {\hspace{1em}}Ours& \textbf{84.1} & 88.5 & 94.0 & \underline{91.8} & \textbf{94.1} & \underline{77.8} & \textbf{50.4} & \textbf{93.3} & \textbf{77.1} & 87.7 & 86.6 & \underline{84.1}\textbf{\textcolor{Maroon}{(+1.7)}}\\
        \bottomrule
    \end{tabular}}
    \caption{Comparison with state-of-the-art methods on MTIL benchmark (Order II) in terms of ``Transfer'', ``Average'', and ``Last'' scores (\%). ``Ours$\dagger$'' and ``Ours'' indicate our method trained on 3k and 1k iterations, respectively. We label the best and second methods with \textbf{bold} and \underline{underline} styles. The top block indicates the upper-bound solutions to adapt the CLIP on each task.}
    \label{tab:suppl3}
\end{table*}

\begin{table*}[t]
	\centering
        \resizebox{0.99\linewidth}{!}{
	\begin{tabular}{cl>{\centering\arraybackslash}p{1cm} >{\centering\arraybackslash}p{1cm}>{\centering\arraybackslash}p{1cm} >{\centering\arraybackslash}p{1cm} >{\centering\arraybackslash}p{1cm} >{\centering\arraybackslash}p{1cm} >{\centering\arraybackslash}p{1cm} >{\centering\arraybackslash}p{1cm} >{\centering\arraybackslash}p{1cm} >{\centering\arraybackslash}p{1cm} >{\centering\arraybackslash}p{1cm} >{\centering\arraybackslash}p{1.5cm}}
        \toprule
           &  {\hspace{1em}}    Method & \makecell[c]{\rotatebox{90}{Cars~\cite{maji2013fine}}} & \makecell[c]{\rotatebox{90}{Food~\cite{fei2004learning}}} & \makecell[c]{\rotatebox{90}{MNIST~\cite{krizhevsky2009learning}}} & \makecell[c]{\rotatebox{90}{OxfordPet~\cite{cimpoi2014describing}}} & \makecell[c]{\rotatebox{90}{Flowers~\cite{helber2019eurosat}}} & \makecell[c]{\rotatebox{90}{SUN397~\cite{nilsback2008automated}}} & \makecell[c]{\rotatebox{90}{Aircraft~\cite{bossard2014food}}} & \makecell[c]{\rotatebox{90}{Caltech101~\cite{deng2012mnist}}} & \makecell[c]{\rotatebox{90}{DTD~\cite{parkhi2012cats}}} & \makecell[c]{\rotatebox{90}{EuroSAT~\cite{krause20133d}}} & \makecell[c]{\rotatebox{90}{CIFAR100~\cite{xiao2010sun}}} & \makecell[c]{{\textit{Average}}} \\
  
        \midrule
        
            \multirow{3}{*}{\rotatebox{90}{CLIP}}& {\hspace{1em}}Zero-shot & 64.7 & 88.5 & 59.4 & 89.0 & 71.0 & 65.2 & 24.3 & 88.4 & 44.6 & 54.9 & 68.2 & 65.3  \\
            & {\hspace{1em}}5-shot Full Fine-tune & 65.4 & 83.3 &96.6&84.9&92.9&71.3&30.6&93.5&65.1&91.7&76.8& 77.5  \\
            & {\hspace{1em}}5-shot Fine-tune Adapter &68.2&87.8&90.4&89.0&94.2&72.5&29.7&90.0&63.9&81.1&75.3 &76.6 \\
            
            \midrule\midrule
            \multirow{7}{*}{\rotatebox{90}{\textbf{Transfer}}} &{\hspace{1em}}Continual-FT &&76.0&\underline{64.6}&67.1&49.7&53.7&8.3&77.9&33.9&23.9&37.1&49.2  \\
            & {\hspace{1em}}LwF \cite{li2017learning} &&64.2&59.1&68.1&38.4&54.9&6.7&78.0&35.5&33.5&47.4 & 48.6  \\
            & {\hspace{1em}}LwF-VR \cite{ding2022don} &&80.1&55.4&77.7&50.4&61.4&9.1&83.5&40.1&31.5&54.8&54.4 \\
            & {\hspace{1em}}WiSE-FT \cite{wortsman2022robust} &&77.3&60.0&76.9&54.2&58.0&11.1&81.8&37.6&31.7&48.1&53.7\\
            & {\hspace{1em}}ZSCL \cite{zheng2023preventing} &&\underline{87.3}&\textbf{64.8}&\underline{85.3}&\underline{67.9}&\underline{64.5}&\textbf{18.9}&\underline{86.6}&\underline{43.6}&\underline{43.2}&\underline{65.7} &\underline{62.8} \\
            \rowcolor{Thistle!20}
            & {\hspace{1em}}Ours &&\textbf{88.8}&59.5&\textbf{89.1}&\textbf{71.2}&\textbf{65.3}&\underline{18.2}&\textbf{87.9}&\textbf{44.2}&\textbf{54.6}&\textbf{68.2}&\textbf{64.7\textcolor{Maroon}{(+1.9)}} \\
            \midrule 
            \multirow{7}{*}{\rotatebox{90}{\textbf{Average}}} &{\hspace{1em}}Continual-FT    & 50.1 & 56.9 & 73.5 & 64.5 & 45.9 & 51.2 & 8.2 & 81.8 & 37.9 & 29.9 & 38.6 & 49.0 \\
            & {\hspace{1em}}LwF \cite{li2017learning} & \underline{64.1} & 55.0 & 79.5 & 69.2 & 55.7 & 58.3 & 10.8 & 81.7 & 41.3 & 39.2 & 47.4 & 54.7  \\
            & {\hspace{1em}}LwF-VR \cite{ding2022don} & 63.3&76.9&71.4&79.1&68.9&65.0&13.4&86.0&45.7&36.3&55.3&60.1\\
            & {\hspace{1em}}WiSE-FT \cite{wortsman2022robust} & 59.3 &64.7&77.4&70.3&51.3&58.6&10.8&84.2&42.0&38.6&49.1&55.1\\
            & {\hspace{1em}}ZSCL \cite{zheng2023preventing} & \textbf{70.0}&\underline{85.0}&\underline{79.8}&\underline{86.1}&\textbf{79.4}&\underline{68.3}&\underline{21.8}&\underline{88.8}&\underline{48.8}&\underline{49.3}&\underline{66.5}&\underline{67.6}\\
            \rowcolor{Thistle!20}
            & {\hspace{1em}}Ours &61.2 &\textbf{87.0}&\textbf{87.3}&\textbf{89.1}&\underline{79.3}&\textbf{68.5}&\textbf{23.4}&\textbf{89.4}&\textbf{49.9}&\textbf{60.8}&\textbf{68.8}& \textbf{69.5}\textbf{\textcolor{Maroon}{(+1.9)}}\\
            \midrule
            \multirow{7}{*}{\rotatebox{90}{\textbf{Last}}} &{\hspace{1em}}Continual-FT &35.2&28.6&58.3&51.2&14.0&46.1&5.3&89.5&47.0&52.9&53.6&42.8 \\
            & {\hspace{1em}}LwF \cite{li2017learning} &57.1&40.1&\underline{84.1}&58.1&50.5&57.6&14.3&87.9&54.7&64.0&47.0&56.8 \\
            & {\hspace{1em}}LwF-VR \cite{ding2022don}&57.3&70.1&72.1&74.6&71.9&65.8&17.4&89.5&60.0&56.0&60.2&63.5 \\
            & {\hspace{1em}}WiSE-FT \cite{wortsman2022robust}&48.1&47.7&66.9&59.8&25.0&56.1&7.4&88.5&52.2&66.8&59.4&51.8\\
            & {\hspace{1em}}ZSCL \cite{zheng2023preventing} &\textbf{67.4}&\underline{82.7}&78.7&\underline{85.7}&\underline{81.3}&\underline{71.2}&\underline{25.0}&\textbf{92.5}&\underline{62.0}&\underline{72.2}&\underline{74.4}&\underline{71.8}\\
            \rowcolor{Thistle!20}
            & {\hspace{1em}}Ours& \underline{59.4}&\textbf{87.0}&\textbf{91.8}&\textbf{89.0}&\textbf{84.1}&\textbf{71.9}&\textbf{29.4}&\underline{91.4}&\textbf{64.2}&\textbf{88.8}&\textbf{75.0}&\textbf{75.7\textcolor{Maroon}{(+3.9)}}\\
        \bottomrule
    \end{tabular}}
    \caption{Comparison with state-of-the-art methods on few-shot MTIL benchmark (Order II) in terms of ``Transfer”, ``Average”, and ``Last” scores (\%). Ours converges in 500 iterations on few-shot. We label the best and second methods with \textbf{bold} and \underline{underline} styles. The top block indicates the upper-bound solutions to adapt the CLIP on each task.
    }
    \label{tab:suppl4-fewshot}
\end{table*}

